\documentclass{article}

\PassOptionsToPackage{numbers, compress}{natbib}
\usepackage[final]{neurips_2022}

\usepackage{microtype}
\usepackage{graphicx}
\usepackage{subfigure}
\usepackage{booktabs}
\usepackage{natbib}
\usepackage{hyperref}

\usepackage{amsmath}
\usepackage{amssymb}
\usepackage{mathtools}
\usepackage{caption}
\usepackage{amsthm}
\usepackage{makecell}
\usepackage{graphicx}
\usepackage{subfigure}
\usepackage{bm}
\usepackage{multirow}
\usepackage{multicol}
\usepackage{hyperref}
\usepackage{url}
\usepackage{optidef}
\usepackage{relsize}
\usepackage{xcolor}
\usepackage{wrapfig,lipsum}

\definecolor{myblue}{rgb}{0, 0.6875, 0.9375}
\definecolor{myorange}{rgb}{0.9451, 0.2980, 0.192157}
\definecolor{mygraygray}{rgb}{0.51465,0.59216,0.6902}
\theoremstyle{plain}
\newtheorem{theorem}{Theorem}[section]

\theoremstyle{definition}

\newtheorem{assumption}[theorem]{Assumption}
\theoremstyle{remark}
\newtheorem{remark}[theorem]{Remark}
\newcommand{\bx}{\bm{\mathrm{x}}}
\newcommand{\bz}{\bm{\mathrm{z}}}
\newcommand{\bu}{\bm{\mathrm{u}}}
\newcommand{\bv}{\bm{\mathrm{v}}}
\newcommand{\bw}{\bm{\mathrm{w}}}
\newcommand{\bA}{\bm{\mathrm{A}}}
\newcommand{\bP}{\bm{\mathrm{P}}}

\definecolor{mygreen}{RGB}{0 205 0}

\hypersetup{
	colorlinks=true,
	urlcolor=magenta,
	citecolor=mygreen,
}

\title{A Closer Look at the Adversarial Robustness of \\ Deep Equilibrium Models}

\author{%
  Zonghan Yang$^1$, Tianyu Pang$^2$, Yang Liu$^{1,3,4}$\thanks{Corresponding author: Yang Liu} \\
  $^1$Department of Computer Science and Technology, Tsinghua University, Beijing, China\\
  $^2$Sea AI Lab, Singapore\\
  $^3$Institute for AI Industry Research (AIR), Tsinghua University, Beijing, China\\
  $^4$Beijing Academy of Artificial Intelligence, Beijing, China\\
  \small{\texttt{yangzh20@mails.tsinghua.edu.cn}, \texttt{tianyupang@sea.com}, \texttt{liuyang2011@tsinghua.edu.cn}}
}

\begin{document}

\maketitle

\vspace{-10pt}
\begin{abstract}
Deep equilibrium models (DEQs) refrain from the traditional layer-stacking paradigm and turn to find the fixed point of a single layer. DEQs have achieved promising performance on different applications with featured memory efficiency. At the same time, the adversarial vulnerability of DEQs raises concerns. Several works propose to certify robustness for monotone DEQs. However, limited efforts are devoted to studying empirical robustness for general DEQs. To this end, we observe that an adversarially trained DEQ requires more forward steps to arrive at the equilibrium state, or even violates its fixed-point structure. Besides, the forward and backward tracks of DEQs are misaligned due to the black-box solvers. These facts cause gradient obfuscation when applying the ready-made attacks to evaluate or adversarially train DEQs. Given this, we develop approaches to estimate the intermediate gradients of DEQs and integrate them into the attacking pipelines. Our approaches facilitate fully white-box evaluations and lead to effective adversarial defense for DEQs. Extensive experiments on CIFAR-10 validate the adversarial robustness of DEQs competitive with deep networks of similar sizes.
\end{abstract}

\vspace{-5pt}
\section{Introduction} \label{section-introduction}
\vspace{-3pt}

Conventional deep networks employ multiple stacked layers to process data in a feedforward manner \cite{he2016deep}. During training, network parameters are optimized by backpropagating loss updates through the consecutive layers \cite{backprop}. Recently, \cite{deq} propose deep equilibrium models (DEQs), whose forward pass involves finding the fixed point (i.e., equilibrium state) of a single layer. With implicit differentiation, the backward pass of DEQs is formulated as another linear fixed-point system. Training DEQs with black-box root solvers only consumes $\mathcal{O}(1)$ memory, which enables DEQs to achieve performance competitive with conventional networks in large-scale applications, including language modelling \cite{deq}, image classification and segmentation \cite{mdeq}, density modelling \cite{impflow,deq-input-opt}, and graph modelling \cite{deq-gnn}.

Considering the fixed point as a local attractor, DEQs are expected to be stable to small input perturbations. However, empirical observations show the opposite that a vanilla DEQ is also vulnerable to adversarial attacks \cite{deq-input-opt}. Along this routine, several works are proposed to investigate the certified robustness for monotone DEQs \cite{mondeq,certiDEQ-1,certiDEQ-2,certiDEQ-3,certiDEQ-4,certiDEQ-5}. Inspired from the monotone operator splitting theories, monotone DEQs are designed with the guarantee of existence and convergence of equilibrium points. However, the layer parameterization of monotone DEQs and the limited scalability of certification methods narrow the scope of these previous studies. On the other hand, \cite{deq-input-opt} explore the adversarial robustness for general DEQs. They incorporate the adversarial generation process into the equilibrium solver to accelerate the PGD attack \cite{madry2018towards}. Nevertheless, the PGD attack is originally designed for deep networks, requiring for end-to-end white-box differentiation. In contrast, DEQs rely on black-box solvers and could obfuscate the gradients used in PGD: as shown in Fig. \ref{fig:figure1-real}-(a), in DEQs trained with different configurations, the intermediate states \textit{always} exhibit higher robustness than the final state under \textit{ready-made} PGD attacks.
Compared to the extensive literature on the adversarial robustness of deep networks \cite{biggio2013evasion,Szegedy2013,Goodfellow2014,Moosavidezfooli2016,kurakin2016adversarial,madry2018towards,zhang2019theoretically,rice2020overfitting,pang2020bag}, much less is known about the adversarial robustness of general DEQs, especially under a well-elaborate white-box setting. This motivates us to disentangle the modules in DEQs and provide a fair evaluation of their robustness.

In this paper, we first summarize the challenges of training robust DEQs (see Sec.~\ref{section-challenges}), including (\textbf{\romannumeral 1}) convergence of the black-box solvers and (\textbf{\romannumeral 2}) misalignment between the forward and backward passes. The off-the-shelf attacks work in a gray-box setting as they have no access to the intermediate states in the forward pass. To thoroughly evaluate the robustness, we propose two methods for intermediate gradient estimation: the first one is iterating adjoint gradient estimations simultaneously in the forward pass, as formally described in Sec.~\ref{sec41}; the second one is estimating intermediate gradients by unrolling, as seen in Sec.~\ref{sec42}. Then in Sec.~\ref{sec5}, we develop approaches to integrate the estimated gradients into the ready-made attacks towards fully white-box adversaries. We also design defense strategies for DEQs to boost their robustness under white-box attacks.

We use PGD-AT to train large-sized and XL-sized DEQs on CIFAR-10. To benchmark their robustness \cite{croce2021robustbench}, the parameter sizes of the DEQs are set to be comparable with ResNet-18 \cite{He2015} and WideResNet-34-10 \cite{zagoruyko2016wide}, respectively. We observe that the adversarially trained DEQs with the exact gradient \cite{deq} require more forward steps to arrive at the equilibrium state, or even violate their fixed-point structures. We also find an intriguing robustness accumulation effect that the intermediate states in the forward pass are more robust under ready-made attacks. These phenomena exhibit gradient obfuscation \cite{athalye2018obfuscated}, which verifies the necessity of intermediate gradient estimation to construct white-box attacks and defense strategies. 
Robustness performance under the white-box evaluation shows that DEQs achieve competitive or stronger adversarial robustness than deep networks of similar parameter amounts. Our investigation sheds light on the pros and cons with respect to the adversarial robustness of DEQs.

\vspace{-5pt}
\section{Background} \label{section-background} 
\vspace{-3pt}
This section includes the background on DEQs and adversarial robustness for deep networks.
\subsection{Deep equilibrium models}

We first briefly introduce the modelling of deep equilibrium models (DEQs) \cite{deq,mdeq}. Consider a $T$-layer weight-tied input-injected neural network:
\begin{equation}
    \bz_{n} = f_\theta(\bz_{n-1}; \bx), ~n=1,\dots,T,
\label{unrolled-ijnn}
\end{equation}
where $\bx \in \mathbb{R}^{l}$ is the input, $\bz_n \in \mathbb{R}^{d}$ is the output of the $n$-th layer, and $\theta$ is the network weights shared across different layers. One can cast the evolution of $\{\bz_n\}$ as a fixed-point iteration process. When $n \to \infty$, $\bz_n$ converges to the fixed point $\bz^*$ which satisfies the equation $\bz^{*} = f_\theta(\bz^*; \bx)$.


Deep equilibrium models rely on the fixed-point equation and leverage a black-box solver to \textit{directly} solve for $\bz^*$ in the forward pass. 
The backward pass of DEQs can also be formulated as a fixed-point iteration process. 
With the loss function $L(\bz^*, y)$ and implicit differentiation, we can compute the gradient with respect to $\theta$ or $\bx$ with 
\begin{equation}
    \frac{\partial L}{\partial (\cdot)} = \left(\frac{\partial f_\theta(\bz^*; \bx)}{\partial (\cdot)}\right) \underbrace{\left( I - \frac{\partial f_\theta(\bz^*; \bx)}{\partial \bz} \right)^{\rm -1} \frac{\partial L(\bz^{*}, y)}{\partial \bz}}_{\bu^*},
\label{exact-gradient}
\end{equation}
where $(\partial \mathbf{a}/\partial \mathbf{b})_{ij} = \partial \mathbf{a}_j / \partial \mathbf{b}_i$ and $\bu^*$ satisfies
\begin{equation}
    \bu^* = \left(\frac{\partial f_\theta(\bz^*; \bx)}{\partial \bz}\right) \bu^* + \frac{\partial L(\bz^{*}, y)}{\partial \bz}.
\label{backward-iter}
\end{equation}
According to Eq. (\ref{backward-iter}), the backward pass can also be executed with a black-box fixed-point solver, and this iteration process is independent of that in the forward pass.

Several techniques have been proposed to improve the training stability of DEQs. \cite{jacobian-deq} propose to regularize the Jacobian matrix in Eq. (\ref{exact-gradient}) during training so that the nonlinear forward system and the backward linear system enjoy appropriate contractivity. \cite{deq-phantom-grad} propose unrolling-based and Neumann-series-based phantom gradients to replace the exact gradient in Eq. (\ref{exact-gradient}) for acceleration. The unrolling-based phantom gradient is defined as
\begin{equation}
    \lambda\sum_{t=1}^{k} \left(\frac{\partial f_\theta\left(\hat{\bz}_{N+t}; \bx\right)}{\partial (\cdot)}\right) \bP_{\lambda, \bz_N}^{(t)} \frac{\partial L(\hat{\bz}_{N+k},y)}{\partial \bz},
\label{unroll-phantom-gradient}
\end{equation}
where 
\begin{equation}
    \bP_{\lambda, \bz_N}^{(t)} = \prod_{s=t+1}^{k} \left(\lambda \frac{\partial f_\theta\left(\hat{\bz}_{N+s}; \bx\right)}{\partial \bz} + (1-\lambda)I\right)\textrm{,}
\end{equation}
\begin{equation}
    \hat{\bz}_{N+t} = (1-\lambda)\hat{\bz}_{N+t-1} + \lambda f_\theta\left(\hat{\bz}_{N+t-1};\bx\right)
\label{unrolled-iterations}
\end{equation}
are the $k$ unrolling steps with $1 \le t \le k$, starting from $\hat{\bz}_N = \bz^*$ returned by the forward solver.

Eq. (\ref{unroll-phantom-gradient}) is calculated by the automatic differentiation framework \cite{paszke2019pytorch} on the computational subgraph in Eq. (\ref{unrolled-iterations}). It is demonstrated that the unrolling-based phantom gradient imposes implicit Jacobian regularization effect to DEQ training \cite{deq-phantom-grad}. DEQs trained by either the exact or the phantom gradients are competitive to deep neural networks in terms of natural accuracy. In our work, we leverage adversarial defense strategies to train DEQs to improve their robustness.

\begin{figure*}[t]
    \begin{subfigure}
        \centering    
        \includegraphics[width=0.48\textwidth]{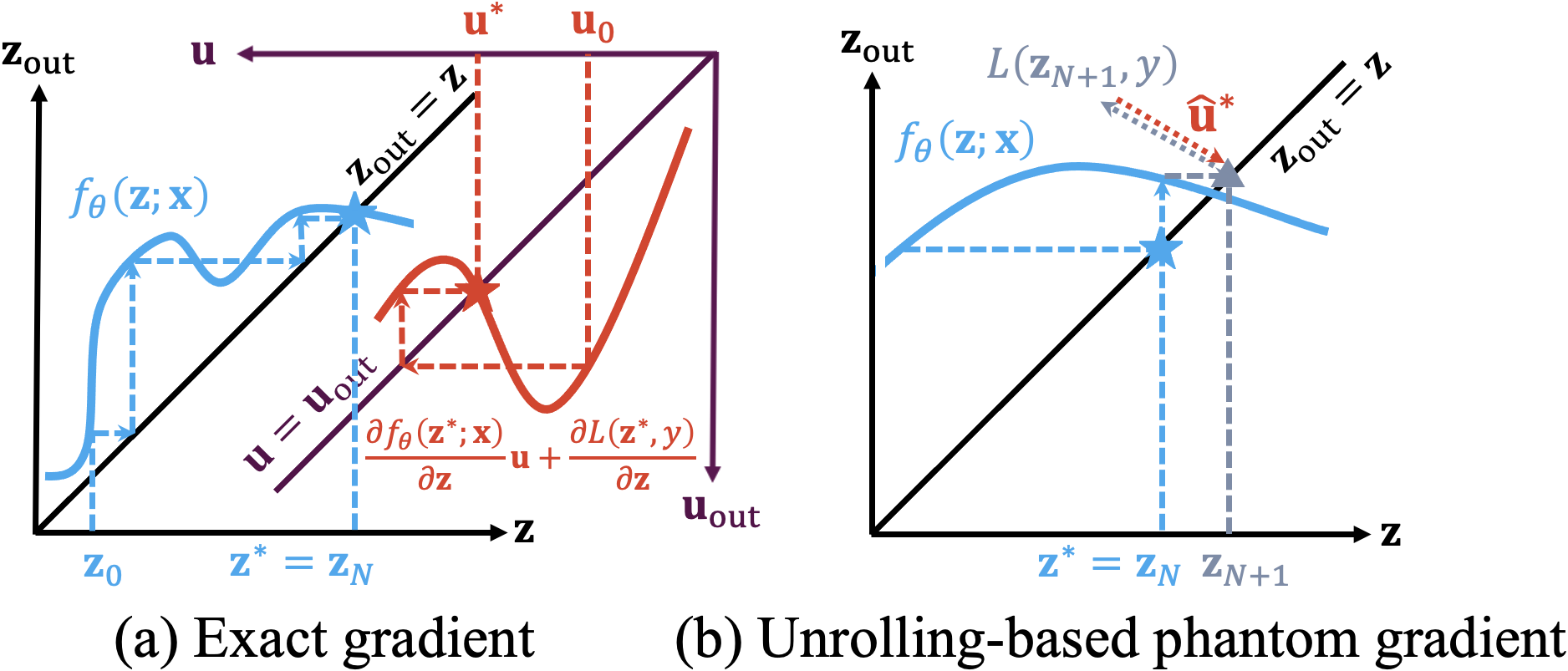}
        \captionsetup{labelformat=empty}
    \end{subfigure}
    \begin{subfigure}
        \centering    
        \includegraphics[width=0.48\textwidth]{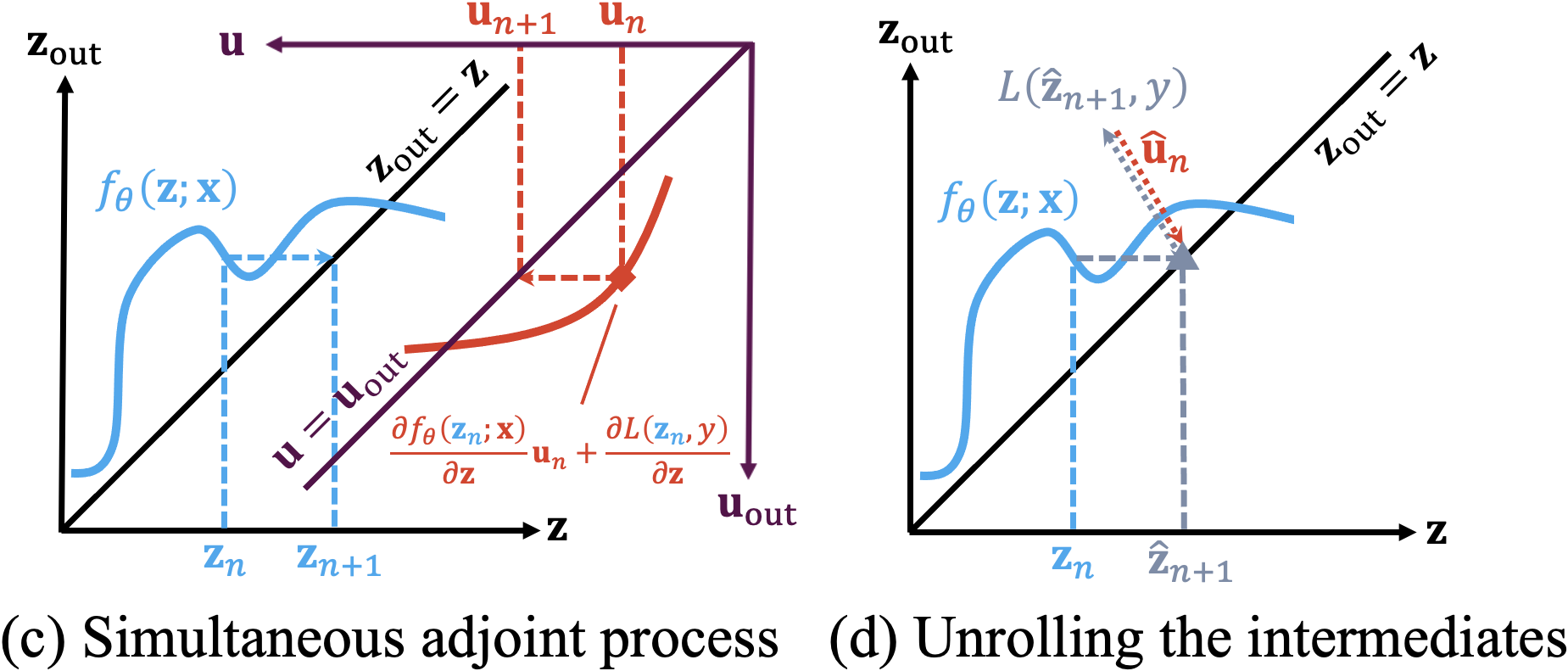}
        \captionsetup{labelformat=empty}
    \end{subfigure}    
    \caption{The gradients proposed for DEQs. (a): the exact gradient \cite{deq} solved by an independent fixed-point iteration process. (b): the unrolling-based phantom gradient \cite{deq-phantom-grad} returned by automatic differentiation on a computational subgraph where the equilibrium state $\bz^*$ is unrolled. (c): simultaneous adjoint process along with the forward iterations described in Sec. \ref{sec41}. (d): unrolling the intermediate states $\bz_n$ for gradient estimation in Sec. \ref{sec42}. We leverage (c) and (d) to estimate intermediate gradients and design fully white-box attacks to evaluate the robustness of DEQs.}
    \label{fig:main-methods}
    \vspace{-15pt}
\end{figure*}

\vspace{-5pt}
\subsection{Adversarial robustness for deep networks}

Much research has been dedicated to adversarial attacks and defenses of deep neural networks. On the one hand, white-box adversarial attack techniques like PGD \cite{madry2018towards} construct adversaries by iteratively perturbing inputs in the gradient ascent direction. The robustness of deep networks is benchmarked by AutoAttack \cite{croce2021robustbench}, which consists of four attacks including two PGD variants with adaptive stepsize and the query-based SQUARE attack \cite{andriushchenko2020square}. On the other hand, adversarial training (AT) \cite{madry2018towards} is one of the most effective defense strategies. By early stopping the training procedure as in \cite{rice2020overfitting}, the primary PGD-AT framework still achieves competitive robustness performance compared with the state-of-the-art defense techniques like TRADES \cite{zhang2019theoretically}. It is worth mentioning that many defense approaches claim robustness improvement by obfuscating gradients, which proves to be a false sense of security under adaptive attacks designed specifically \cite{athalye2018obfuscated}.
In our work, we train DEQs with PGD-AT and investigate their adversarial robustness by designing customized defenses and adaptive attacks.

\vspace{-5pt}
\section{Challenges for robust general DEQs} \label{section-challenges}
This section describes the challenges encountered when we aim to train robust general DEQs.

\textbf{Misalignment between forward \& backward passes.}\label{subsection:challenges-misalignment} The central idea of DEQs is \textit{directly} solving for the equilibrium state $\bz^*$ and differentiating through the fixed point equation $\bz^* = f_\theta(\bz^*; \bx)$ for efficient forward and backward passes. Fig. \ref{fig:main-methods}-(a) sketches the calculation of the exact gradient \cite{deq}. Independent from the forward iterations (the \textbf{\textcolor{myblue}{blue}} curve), the exact gradient is acquired by solving for a linear fixed-point system that only depends on the equilibrium state $\bz^*$ (the \textbf{\textcolor{myorange}{orange}} curve). Fig. \ref{fig:main-methods}-(b) shows the calculation of the unrolling-based phantom gradient \cite{deq-phantom-grad}. $\bz^*$ as the final state in the forward pass is unrolled (the \textbf{\textcolor{mygraygray}{gray}} iteration), and the gradient is obtained from the automatic differentiation on the loss function. However, when iterating the gradient computations, the intermediate states $\{\bz_n\}$ in the forward pass are bypassed by both methods. The misalignment between the forward and backward tracks results in a gray-box setting for the ready-made attacks.

\textbf{Convergence of the black-box solvers.} In contrast with monotone DEQs, there is no guarantee for the existence and convergence of the equilibrium states in general DEQs. It is thus unknown whether the black-box solvers in DEQs still converge to equilibrium states under input perturbations. Adversarial training also adds the concern on equilibrium convergence. The well-known effect of adversarial training for deep networks is the trade-off between robustness and accuracy \cite{Su_2018_ECCV,tsipras2018robustness,zhang2019theoretically,raghunathan2020understanding}. A similar drop in standard accuracy (from $78$\% to $55$\%) is also observed for tiny-sized adversarially-trained DEQs \cite{deq-input-opt}. The robustness-accuracy trade-off brings training instability for general DEQs, which may take more iterations in the solvers for equilibrium convergence, or even violate their fixed-point structures. Finally, the robustness comparison is still under-explored between large-sized general DEQs and deep networks with similar parameter counts.

\vspace{-5pt}
\section{On intermediate gradient estimation} \label{section-grad-intermediate} 
\vspace{-5pt}



As the forward and backward tracks in DEQs are misaligned, the intermediate states in the forward pass are inaccessible to off-the-shelf attacks, which causes gradient obfuscation and results in false positive robustness. Therefore, it is necessary to estimate the intermediate gradients. With the integration of the estimated gradients, the attacks can validate the robustness of DEQs in a fully white-box setting. In this section, we propose two methods for intermediate gradient estimation.

\vspace{-5pt}
\subsection{Simultaneous adjoint in the forward pass}
\label{sec41}
Inspired by the adjoint process in neural ODE models \cite{neuralode}, we propose the adjoint process for intermediate gradient estimation in DEQs. 
The adjoint process in neural ODE models is characterized by an adjoint ODE \cite{ode-adjoint}. For DEQs, we propose to iterate the updates of adjoint states subject to $\bz_n$ in the forward pass. We investigate the simultaneous adjoint with Broyden's method \cite{Broyden} as the forward solver. 
In the forward pass, Broyden's method updates the intermediate state $\bz_n$ based on the residual $g_\theta(\bz_n;\bx) = f_\theta(\bz_n;\bx) - \bz_n$ and $B_n$, the low-rank approximation of the Jacobian inverse:
\begin{align}
    \bz_{n+1} &= \bz_n - \alpha B_{n}g_\theta(\bz_n;\bx), ~\bz_0 = \mathbf{0}\\
    B_{n+1} &= B_n + \frac{\Delta\bz_{n+1} - B_n \Delta g_{n+1}}{\Delta\bz_{n+1}^{\rm T}B_n \Delta g_{n+1}} \Delta\bz_{n+1}^{\rm T}B_n,
\end{align}
where $0 \le n \le N-1$, $B_0 = -I$, $\Delta\bz_{n+1} = \bz_{n+1} - \bz_n$, $\Delta g_{n+1} = g_\theta(\bz_{n+1};\bx) - g_\theta(\bz_{n};\bx)$, and $\alpha$ is the step size. 
To maintain a simultaneous adjoint, we start from $\bu_0 = \mathbf{0}$ and use Broyden's method to solve Eq. (\ref{backward-iter}). Similar with the residual function $g_\theta(\cdot;\bx)$ for $\bz_n$, the fixed-point equation in Eq. (\ref{backward-iter}) defines the residual of the adjoint state. However, we propose to replace the $\bz^*$ in Eq. (\ref{backward-iter}) by $\bz_n$, and integrate the approximated Jacobian inverse $B_n$ to force the alignment of the adjoint state update:
\begin{align}
    \bv_n &= \left(\frac{\partial f_\theta(\bz_n; \bx)}{\partial \bz}\right) \bu_n + \frac{\partial L(\bz_n, y)}{\partial \bz} - \bu_n, \label{adjoint-residual-v} \\ 
    \bu_{n+1} &= \bu_n - \beta B_{n} \bv_n, \label{adjoint-update}
\end{align}
where $\bv_n$ is the residual at iteration $n$, $\bu_n$ is the updated adjoint state, and $\beta>0$ is the step size. 

We use the following surrogate gradients to construct attacks on the intermediate state $\bz_n$:
\begin{equation}
    \left[\widetilde{\frac{\partial L}{\partial x}}\right]_n = \left( \frac{\partial f_\theta(\bz_n; \bx)}{\partial \bx} \right) \bu_n.
\label{sm-adjoint-gradient}
\end{equation}
An illustration for the simultaneous adjoint process is shown in Fig. \ref{fig:main-methods}-(c). In the following, we refer to this method as \textit{simultaneous adjoint} when constructing intermediate state attacks in Sec \ref{section:white-box-attacks}.

\begin{remark}
    We show in Appendix \ref{app:theorem-proof} that under mild assumptions, the $\{\bu_n\}$ converges to $\bu^*$ when $0 < \beta < 1$ in Eq. (\ref{adjoint-update}). However in practice, we do \textit{not} require the convergence of $\bu_n$ as we only use them in Eq. (\ref{sm-adjoint-gradient}) as gradient \textit{estimations} to construct \textit{intermediate attacks}.
\end{remark}

\begin{remark}
Similar with the update of $\bu_n$ in our approach, \cite{deq-input-opt} propose augmented DEQs as an integration of the iterative updating process of $\bz$, $\bu$, and $\bx$ as a whole:
\begin{equation}
    F\left(
    \begin{bmatrix}
        \bz_{n} \\ \bu_{n} \\ \bx_{n} \\
    \end{bmatrix}
    \right)
    = 
    \begin{bmatrix}
        f_\theta(\bz_{n}; \bx_{n}) \\  
        \left(\frac{\partial f_\theta(\bz_n; \bx_n)}{\partial \bz}\right) \bu_{n} \!+\! \frac{\partial L(\bz_n, y)}{\partial \bz} \\ 
        \bx_{n} - \left(\frac{\partial f_\theta(\bz_n; \bx_{n})}{\partial \bx} \right) \bu_{n} \\
    \end{bmatrix}
\end{equation}
The augmented DEQs leverage a black-box solver (e.g., Broyden's method) to find the equilibrium of the whole state $[\bz^*, \bu^*, \bx^*]^{\rm T}$. However, several cross-terms exist in the joint Jacobian due to the coupling of the three iteration processes, which further hinders the convergence. In contrast, the simultaneous adjoint update in Eq. (\ref{adjoint-update}) does not include the update of $\bx$. Furthermore, we reuse the Jacobian inverse approximation matrix $B_n$ in the update of $\bu_n$, which is easy to implement. Because of the disentanglement of $\bz_n$ and $\bu_n$ in the joint Jacobian, our method also enjoys better efficiency and flexibility as one can early exit the adjoint process without affecting the updates of $\bz_n$'s.    
\end{remark}

\begin{remark}
Concurrent work \cite{SHINE} also explores the idea of sharing approximated Jacobian inverse $B_n$ in bi-level optimization problems. While their motivation is to accelerate DEQ training, we use the adjoint states to construct gradient estimation and facilitate white-box attacks. We also compare our work with theirs in terms of intermediate state attacks; See Appendix \ref{app:adjoint-broyden} for details.
\end{remark}

\subsection{Unrolling the intermediate states} 
\label{sec42}

We also propose to estimate the gradient at the state $\bz_n$ by unrolling. Depicted in Fig. \ref{fig:main-methods}-(d), $\bz_n$ is involved in an artificially constructed computational graph. We can thus estimate the intermediate gradient by backpropagation with automatic differentiation. Formally, applying Eq. (\ref{unroll-phantom-gradient}) to $\bz_n$ yields
\begin{equation}
    \left[\widehat{\frac{\partial L}{\partial \bx}}\right]_n^{(k)} = \bA^{(k)}_{\lambda, \bz_n} \frac{\partial L(\hat{\bz}_{n+k},y)}{\partial \bz},
\label{unroll-intermediate-gradient}    
\end{equation}
where
\begin{align}
    \bA^{(k)}_{\lambda, \bz_n} &= \lambda\sum_{t=0}^{k-1} \left(\frac{\partial f_\theta(\hat{\bz}_{n+t}; \bx)}{\partial \bx} \right) \bP_{\lambda, \bz_n}^{(k)}\label{eq:unroll-intermediate-gradient-with-k}, \\
    \bP_{\lambda, \bz_n}^{(k)} &= \prod_{s=t+1}^{k-1} \left(\lambda \frac{\partial f_\theta(\hat{\bz}_{n+s}; \bx)}{\partial \bz} + (1-\lambda)I\right),
\end{align}
and the state sequence $\hat{\bz}_{n}, \hat{\bz}_{n+1}, \cdots, \hat{\bz}_{n+k}$ represents the damped unrolling iteration:
\begin{equation}
    \hat{\bz}_{n+t} = (1-\lambda)\hat{\bz}_{n+t-1} + \lambda f_\theta(\hat{\bz}_{n+t-1}; \bx),
\end{equation}
with $t=1, 2, \cdots, k$ and $\hat{\bz}_{n} = \bz_n$. While Eq. (\ref{unroll-phantom-gradient}) is proposed as an approximation of the exact gradient, we unroll the states $\bz_n$ for intermediate gradient estimation.
Similar to the case of Eq. (\ref{sm-adjoint-gradient}), we use Eq. (\ref{unroll-intermediate-gradient}) as estimation to design intermediate attacks for DEQs. We refer to this method as \textit{unrolled intermediates} in the following when incorporating Eq. (\ref{unroll-intermediate-gradient}) into the white-box attacks.



\section{White-box attacks and defenses for DEQs}
\label{sec5}
This section describes different types of white-box attacks and defense strategies for DEQs. 

\vspace{-5pt}
\subsection{White-box attacks for DEQs} \label{section:white-box-attacks}
\vspace{-5pt}

The existing attacks leverage the gradients calculated at the \textit{final} state outputted by the forward solver. Based on the surrogate intermediate gradients in Eq. (\ref{sm-adjoint-gradient}) or Eq. (\ref{unroll-intermediate-gradient}), we can involve the $\bz_n$ in the forward pass into the construction of adversaries. A direct white-box approach is to use the estimated gradient at an \textit{early} state $\bz_n$ as an alternative for input perturbations. Another simple yet effective method is to average the intermediate gradients as the gradient ensemble for attacks. For example, the average of all intermediate gradients along the simultaneous adjoint process is given by
\begin{equation}
    \mathlarger{\sum}_{n} \left[\widetilde{\frac{\partial L}{\partial \bx}}\right]_n = \mathlarger{\sum}_{n} \left( \frac{\partial f_\theta(\bz_n; \bx)}{\partial \bx} \right) \bu_n.
\label{ensemble-attack}
\end{equation}
The gradient ensemble can be viewed as the fusion of all perturbation directions indicated by all $\bz_n$'s.  

\vspace{-5pt}
\subsection{Defenses with intermediate states} \label{section:white-box-defenses}
\vspace{-5pt}

In addition to the final state $\bz^*$, the unused intermediate states can be leveraged as well for the defenses of DEQs. A simple yet effective defense strategy is to \textit{early exit} the forward solver during inference. 
We can evaluate the robustness of DEQs with the early state $\bz_n$ in the forward pass, as $\bz_n$ and $\bz^*$ have the same shape. We determine the optimal timing for early exit by selecting the top robustness performance of all $\bz_n$'s on the development set under the ready-made PGD-10 attack. 

The input-injected neural network provides an interpretation for the early-state defense. From Eq. (\ref{unrolled-ijnn}), the distortion of $\bz_n$ comes from both the perturbed $\bz_{n-1}$ and the biased transformation $f_\theta(\cdot; \bx\!+\!\Delta\bx)$. By early exiting the forward process, one obtains a less distorted intermediate state.

Another defense strategy for DEQs is leveraging the \textit{ensemble} of intermediate states. Similar with Eq. (\ref{ensemble-attack}), we average the intermediate states $\{\bz_n\}$ to defend against attacks. Instead of early stopping, the intermediate state ensemble exploits the state representations at all iterations in the forward solver.

While the proposed defense techniques leverage the intermediate states, they still require only $\mathcal{O}(1)$ memory. For early state defense, we determine the optimal time to early exit the solver on the development set offline for once and then fix the early exit step during testing. For ensemble state defense, we maintain an accumulator to sum up $\{\bz_n\}$ along the forward pass without storing them.

\begin{figure}[t]
    \centering
    \includegraphics[scale=0.42]{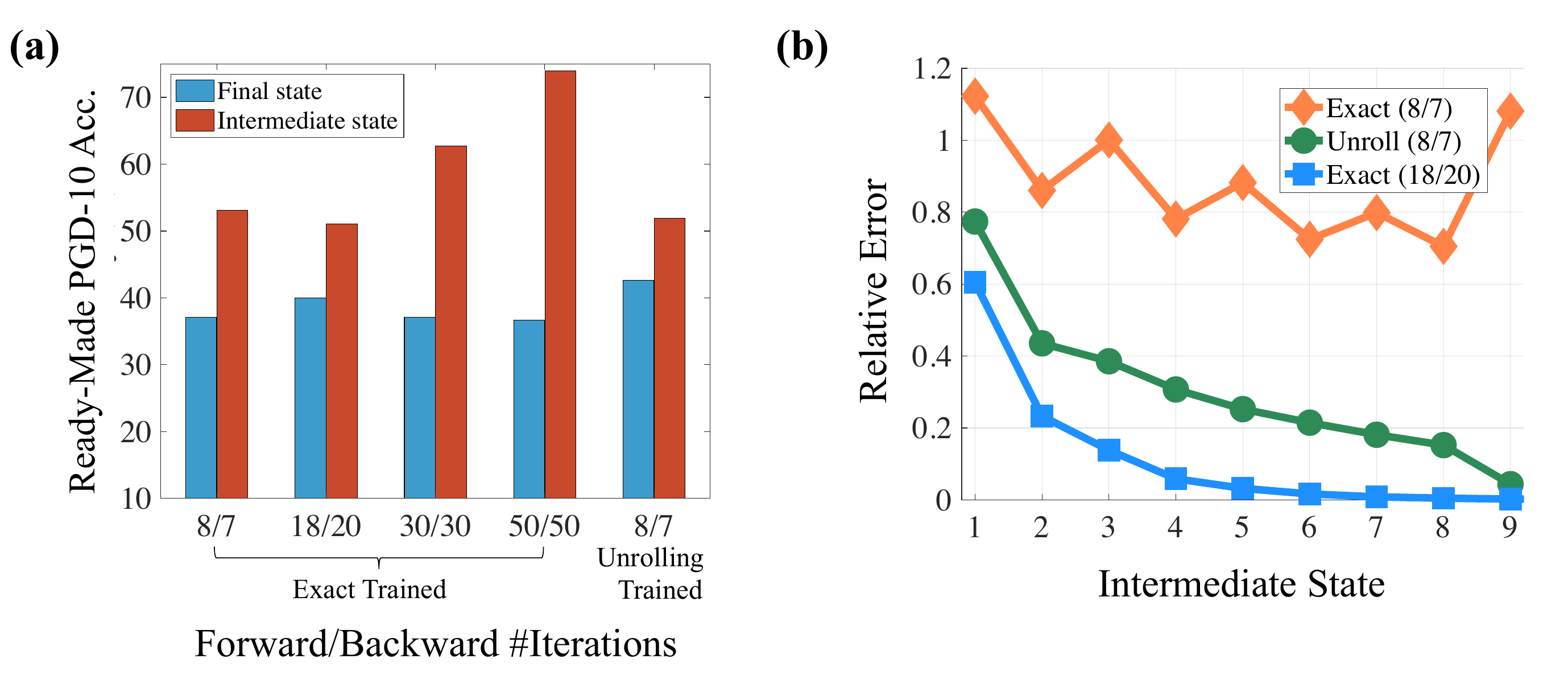}
    \vspace{-0.1cm}
    \caption{Challenges in benchmarking adversarial robustness of DEQs. (a) Gradient obfuscation issues arise in the DEQs trained with different configurations. With different iteration settings in the DEQ solver or different gradient formulations, the intermediate state \textit{always} exhibit higher robustness than the final state under ready-made PGD-10 attack. (b) Exact-trained DEQ with small iterations violate the fixed-point structure and require more iterations to retain it (analyzed in Sec. \ref{sec:violation-of-fps}). Both observations motivate us to design adaptive attacks for white-box robustness evaluation for DEQs.}
    \label{fig:figure1-real}
    \vspace{-0.5cm}
\end{figure}

\vspace{-5pt}
\section{Experiments} \label{section-experiments}

Following the settings in \cite{jacobian-deq}, we experiment with the large-sized DEQ with its parameter count similar to ResNet-18 \cite{he2016deep}. We also experiment with an XL-sized DEQ with its parameter count similar to WideResNet-34-10 \cite{zagoruyko2016wide} to enable a fair comparison with the empirical robustness of the deep networks. The detailed experimental settings are listed in Appendix \ref{app:experimental-settings}.
We first train DEQs on CIFAR-10 \cite{Krizhevsky2012} with the PGD-AT framework \cite{madry2018towards}, then test the adaptive attacks and defense strategies proposed in Sec. \ref{sec5} on the adversarially-trained DEQs. We refer to a DEQ as ``exact-trained" when using the exact gradient, and ``unrolling-trained" when using the unrolling-based phantom gradient in the PGD-AT framework to generate adversaries and optimize for model parameters.
Unless specified, all DEQs are adversarially trained in this paper. During training, we use $10$-step PGD with the step size of $2/255$ to generate adversaries within the range of $\ell_\infty=8/255$. For the specific type of attacks, we use PGD and AutoAttack (AA) \cite{croce2020reliable} to instantiate the white-box attacks in Sec. \ref{section:white-box-attacks}.

\vspace{-5pt}
\subsection{The retention of the fixed-point structure}\label{sec:violation-of-fps}
\vspace{-5pt}

We start with the observation on the fixed-point structure. Shown in Fig. \ref{fig:figure1-real}-(b), the lines illustrate the relative error $\|f_\theta(\bz_n;\bx) - \bz_n\|_2 / \|f_\theta(\bz_n;\bx)\|_2$ for each $\bz_n$ \footnote{The $9$-th intermediate state comes from the implementation in DEQs. In the exact-trained DEQ, $\bz_9 = f_\theta(\bz_8; \bx)$. In the unrolling-trained DEQ, $\bz_9$ is the final state after unrolling $\bz_8$. For the exact-trained DEQ with 18 forward / 20 backward iterations, we only plot the first $9$ states for an easy comparison.}. We find that for the exact-trained DEQ with small iteration settings (8 forward / 7 backward iterations), all the relative errors are larger than $0.75$, i.e., the forward solver in the DEQ fails to converge to an equilibrium. Such phenomenon reflect the challenge on the convergence of the black-box solvers in DEQs mentioned in Section \ref{section-challenges}. It is shown that a larger iteration setting is required (18 forward / 20 backward iterations) for exact-trained DEQs to retain the fixed-point structure. In contrast, we find the small iteration setting (8/7) is enough for the unrolling-trained DEQ to retain the fixed-point structure.

\begin{wraptable}{r}{0.5\textwidth}
    \vspace{-10pt}
    \caption{Performance (\%) of the exact-trained DEQ-Large with the small (8/7) solver iterations under different attacks. The high accuracy under PGD-10 with the exact gradient is deteriorated using the unrolling-based phantom gradient. Leveraging the query-based SQUARE leads to even lower accuracy. These observations indicate that the DEQ with violated fixed-point structure suffers severe robustness degradation.}
    \label{tab:robustness-exact-grad}
    \centering
     \resizebox{0.5\textwidth}{!}{\begin{tabular}{ccccc}
        \toprule
        Gradient & Clean & PGD-10 & PGD-1000 & SQUARE \\
        \midrule
        Exact & \multirow{2}{*}{78.24} & 79.97 & 80.10 & \multirow{2}{*}{5.95} \\
        Unrolling & ~ & 37.07 & 36.39 & ~ \\
        \bottomrule
    \end{tabular}}
    \vspace{-10pt}
\end{wraptable}

It is necessary to retain the fixed-point structure, otherwise leading to gradient obfuscation issues. As is derived from the implicit differentiation on the fixed-point equation $\bz^* = f_\theta(\bz^*; \bx)$, the exact gradient in Eq. (\ref{exact-gradient}) becomes inexact when the equilibrium point $\bz^*$ is not reached. Table \ref{tab:robustness-exact-grad} shows the empirical performance of the exact-trained DEQ under the small (8/7) iteration setting. The severe performance degradation under alternative gradient formulations as well as the SQUARE attack also indicates gradient obfuscation, as suggested in \cite{athalye2018obfuscated} and \cite{carlini2019evaluating}. 

\looseness=-1 While large iterations for the exact-trained DEQs keep the fixed-point structure, it inevitably slows down the training speed (detailed in Appendix \ref{appendix-f-2}). For the exact-trained DEQs with the small (8/7) iteration setting, we have also tried with varied Jacobian regularization weights to impose stricter Lipschitz constraints during training, but found the DEQ solver still diverged. We have also analyzed the instability by tracing the variation of Lipschitz constant during the adversarial training of DEQs; See Appendix \ref{app:jac-reg} for details. By comparison, the unrolling-trained DEQ requires fewer iterations in the forward solver to converge. According to the green line in Fig. \ref{fig:figure1-real}-(b), the relative errors become lower consequently and reach $0.04$ at the final state. The results coincide with \cite{deq-phantom-grad} that the unrolling-based phantom gradient invokes implicit Jacobian regularization during training.
\begin{table*}[t]
    \caption{Performance (\%) of the unrolling-trained DEQ-Large with the small (8/7) iteration setting and the exact-trained DEQ-Large with the large (50/50) iteration setting under PGD-10. The ``final" rows and columns represent the original DEQ output and the ready-made attacks at the final state. The ``early" rows indicate early state defense, and the ``intermediate" columns indicate the performance of the strongest intermediate attacks. The rows and the columns of ``ensemble" demonstrate the ensemble defense and the white-box attacks based on gradient ensemble. Under the (\underline{underlined}) strongest attacks, the ensemble defense achieves the best robustness performance (\textbf{in bold}).}
    \vspace{-0.35cm}
    \label{tab:train-unroll-train-exact-robust-acc-pgd}
    \begin{center}
    \resizebox{\textwidth}{!}{\begin{tabular}{rcccccccc}
        \toprule
        \multirow{2}{*}{Training Configurations} & \multirow{2}{*}{Defense} & \multirow{2}{*}{Clean} & \multicolumn{3}{c}{Simultaneous Adjoint} & \multicolumn{3}{c}{Unrolled Intermediates} \\
        \cmidrule{4-9}
        ~ & ~ & ~ & Final & Intermediate & Ensemble & Final & Intermediate & Ensemble \\
        \midrule
        \multirow{3}{*}{(8/7) Unrolling-Trained} & Final & 78.03 & 49.81 & 59.49 & 54.91 & \underline{42.67} & 62.24 & 51.52 \\
        ~ & Early & 79.57 & 54.90 & 39.19 & 42.76 & 51.90 & \underline{29.38} & 34.20 \\
        ~ & Ensemble & 79.67 & 51.52 & 52.43 & 49.47 & 49.02 & 55.10 & \underline{\textbf{47.12}} \\
        \midrule
        \multirow{3}{*}{(50/50) Exact-Trained} & Final & 73.51 & 37.77 & 70.52 & 43.70 & \underline{36.70} & 69.29 & 48.08 \\
        ~ & Early & 86.98 & 75.25 & \underline{12.44} & 40.12 & 73.93 & 18.24 & 26.22 \\
        ~ & Ensemble & 75.12 & 40.20 & 72.41 & 45.06 & \underline{\textbf{39.18}} & 68.83 & 49.10 \\        
        \bottomrule
    \end{tabular}}
    \end{center}
    \vspace{-0.5cm}
\end{table*}

\vspace{-0.2cm}
\subsection{Robustness of DEQs under white-box attacks} \label{sec:compare-exact-unrolling-under-white-attacks}

Intriguingly, we discover the robustness accumulation effect in both the exact-trained and the unrolling-trained DEQs. We plot the highest robustness under the ready-made PGD-10 among all the intermediate states in Fig. \ref{fig:figure1-real}-(a), with comparison to the final state robustness. It is shown that the intermediate states always exhibit much higher robustness. The accumulated robustness comes from gradient obfuscation, as the ready-made attacks fail to ``directly" attack the intermediate states due to misaligned gradients. This resonates with the first challenge in Sec. \ref{subsection:challenges-misalignment}, and similar results are observed as well in adversarially-trained neural ODEs: the large error tolerance from the ODE solvers with adaptive step sizes allows gradient masking after adversarial training \cite{neuralodegradobfuscation}.

The exact-trained DEQs, as we have discussed in Sec. \ref{sec:violation-of-fps}, require larger iterations in the solver. However, it is noticed in Fig. \ref{fig:figure1-real}-(a) that the larger the iteration is in the exact-trained DEQs, the more robust the intermediate states are under ready-made PGD-10. On the contrary, the (8/7) unrolling-trained DEQ still achieves the highest robustness at the final state. To benchmark the white-box robustness, in this section, we compare the (50/50) exact-trained DEQ-Large with the (8/7) unrolling-trained one \footnote{The robustness of the (8/7) exact-trained DEQ-Large is much lower than its unrolling-trained counterpart because of the violated fixed-point structure, as shown in Sec. \ref{sec:violation-of-fps} and Table \ref{tab:robustness-exact-grad}}.
We integrate the estimated intermediate gradients in Sec. \ref{section-grad-intermediate} as different alternatives in PGD-10 for white-box evaluation. Among all the attack candidates based on intermediate gradients, we select the one that leads to the largest robustness deterioration on the early-state defense in Sec. \ref{section:white-box-defenses} and report the results in Table \ref{tab:train-unroll-train-exact-robust-acc-pgd}. Ablation studies on the performance of the attack candidates with estimated gradients at different intermediates can be found in Sec. \ref{section:ablation-different-intermediates}. We also include the report of memory usage for the defense strategies in Appendix \ref{appendix-g-memory-concern}, and the running time complexity analysis for the white-box attacks in Appendix \ref{appendix-k-time-complexity}.

\looseness=-1 Shown in Table \ref{tab:train-unroll-train-exact-robust-acc-pgd}, for the unrolling-trained DEQ, unrolling the intermediate states results in the strongest attack to the final state and early state defenses. While the robustness accuracies under final and intermediate attacks are improved and better balanced with the ensemble state defense, the ensemble attack leads to the largest performance drop in this case, arriving at the overall white-box robustness of $47.12\%$. The estimated intermediate gradients based on simultaneous adjoint process also shows significant attack performance for the exact-trained DEQ with large solver iterations. After maximizing the minimum robustness under all attacks across all defense techniques, the overall white-box robustness is $39.18\%$. In addition, all attacks significantly deteriorate the robustness of the DEQs without adversarial training, indicating that the attacks leveraged in Table \ref{tab:train-unroll-train-exact-robust-acc-pgd} are reliably strong (detailed in Sec. \ref{appendix-vanilla-deq}). Considering the superior robustness of the unrolling-trained DEQ as well as its training efficiency, we proceed to experiment with unrolling-trained DEQs for further evaluation.



\vspace{-5pt}
\subsection{Comparison between DEQs and deep networks}
\vspace{-2pt}

\begin{table*}[t]
    \caption{Performance (\%) of the unrolling-trained DEQs under \textbf{PGD-10}/\textbf{AutoAttack}. The rows and the columns represent the same meanings as those in Table \ref{tab:train-unroll-train-exact-robust-acc-pgd}. Under the (\underline{underlined}) strongest attacks, the ensemble defense achieves the best robustness performance (\textbf{in bold}).}
    \vspace{-0.35cm}
    \label{tab:train-unroll-robust-acc-pgd-aa}
    \begin{center}
    \resizebox{\textwidth}{!}{\begin{tabular}{ccccccccc}
        \toprule
        \multirow{2}{*}{Arch.} & \multirow{2}{*}{Defense} & \multirow{2}{*}{Clean} & \multicolumn{3}{c}{Simultaneous Adjoint (PGD/AA)} & \multicolumn{3}{c}{Unrolled Intermediates (PGD/AA)} \\
        \cmidrule{4-9}
        ~ & ~ & ~ & Final & Intermediate & Ensemble & Final & Intermediate & Ensemble \\
        \midrule
        \multirow{3}{*}{Large} & Final & 78.03 & 49.81/51.48 & 59.49/61.95 & 54.91/52.95 & \underline{42.67}/\underline{37.27} & 62.24/65.53 & 51.52/49.66 \\
        ~ & Early & 79.57 & 54.90/61.12 & 39.19/55.47 & 42.76/56.84 & 51.90/56.86 & \underline{29.38}/\underline{25.41} & 34.20/49.74 \\
        ~ & Ensemble & 79.67 & 51.52/56.06 & 52.43/58.69 & 49.47/55.02 & 49.02/50.45 & 55.10/58.63 & \underline{\textbf{47.12}}/\underline{\textbf{48.37}} \\
        \midrule
        \multirow{3}{*}{XL} & Final & 82.92 & 55.80/58.21 & 55.80/58.21 & 67.30/64.24 & \underline{48.58}/\underline{43.97} & 65.94/72.76 & 58.23/69.83 \\
        ~ & Early & 80.12 & 51.40/58.92 & 51.40/58.92 & 60.78/62.98 & 52.08/\underline{56.58} & 55.70/62.67 & \underline{48.88}/62.87 \\
        ~ & Ensemble & 81.17 & 52.87/58.08 & 52.87/58.08 & 61.40/62.23 & \underline{\textbf{51.70}}/\underline{\textbf{54.09}} & 59.71/66.90 & 53.23/56.45 \\
        \bottomrule
    \end{tabular}}
    \end{center}
    \vspace{-0.6cm}
\end{table*}

In this section, we further provide a thorough evaluation by benchmarking the white-box robustness performance of the unrolling-trained DEQ-Large and DEQ-XL under both PGD-10 and AutoAttack.

Table \ref{tab:train-unroll-robust-acc-pgd-aa} shows the robustness performance of the unrolling-trained DEQ-Large and DEQ-XL. According to the results, the gradient ensemble attacks are more effective in defeating the early-state defense than the final-state defense. The ensemble attack is the most threatening on the ensemble defense in DEQ-Large. The attack with the gradient at the final state leads to the most substantial performance drop on the ensemble defense in DEQ-XL.

\begin{wraptable}{r}{0.5\textwidth}
    \vspace{-0.4cm}
    \caption{The comparison of robustness performance (\%) between the DEQs with the ensemble defense and the deep networks of similar sizes. For the DEQs, the weakest robustness under all attacks in Table \ref{tab:train-unroll-robust-acc-pgd-aa} is reported. For the deep networks, we report the results in \cite{pang2020bag}.}
    \vspace{-0.1cm}
    \label{tab:deq-dnn-comparison}
    \centering
    \resizebox{0.5\textwidth}{!}{\begin{tabular}{ccccc}
    \toprule
        Arch. & Clean & PGD-10 & AA & \#Params \\
    \midrule
        ResNet-18 & 82.52 & 53.58 & 48.51 & 10M \\
        DEQ-Large & 79.67 & 47.12 & 48.37 & 10M \\
    \midrule
        WRN-34-10 & 86.07 & 56.60 & 52.19 & 48M \\
        DEQ-XL    & 81.17 & 51.70 & 54.09 & 48M \\
    \bottomrule
    \end{tabular}}
    \vspace{-0.4cm}
\end{wraptable}

In Table \ref{tab:train-unroll-robust-acc-pgd-aa}, we find that the PGD-10 attack brings more significant performance drops than AutoAttack does in many settings. The results differ from the case in the robustness of deep networks \cite{croce2020reliable}. The phenomenon originates from the difference between intermediate-state attacks and alternative defense strategies. AutoAttack will overfit to the provided gradients at the intermediate or the averaged states, thus generating less threatening adversaries on the defenses based on other states. In addition, the intermediate gradients can also be inaccurate, as they only serve as approximations.

The minimum robustness under all types of attacks represents the robustness of a defense strategy. We thus take the most robust defenses for DEQ-Large and DEQ-XL, and compare them with the deep networks of similar parameter counts. Shown in Table \ref{tab:deq-dnn-comparison}, the empirical robustness of the DEQs is competitive with or even slightly higher than that of the ResNet-18 and WRN-34-10 models with the PGD-AT framework, respectively.

\vspace{-5pt}
\subsection{Ablation study on different intermediate gradients}\label{section:ablation-different-intermediates}
\vspace{-2pt}

\begin{figure*}
    \centering
    \includegraphics[scale=0.42]{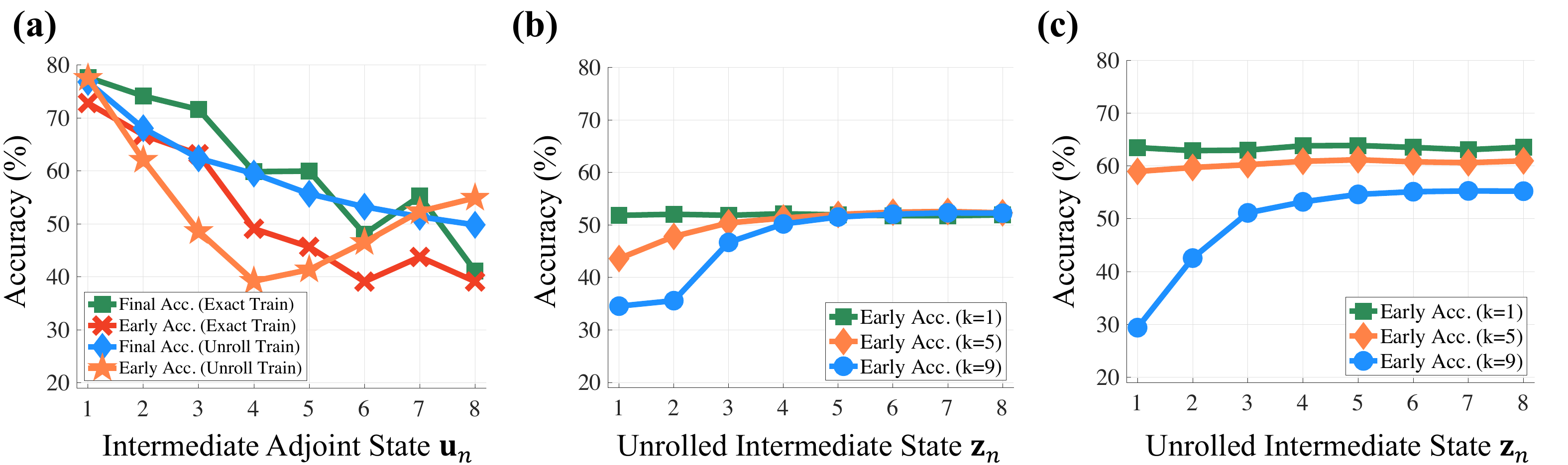}
    \caption{Ablation study on the gradient estimations at different intermediate states.
    (a) Different robustness performance under PGD-10 with different intermediate adjoint states $\bu_n$ as the surrogate gradient. For the approach in Sec. \ref{sec41}, $\bu_4$ leads to the largest robustness drop in the early state $\bz_3$ in the unrolling-based DEQ. (b) and (c): Different unrolled intermediates $\bz_n$ with different $k$'s in Eq. (\ref{eq:unroll-intermediate-gradient-with-k}). the $\lambda$ in Eq. (\ref{unroll-intermediate-gradient}) is set as $0.5$ in (b) and $1$ in (c). For the method in Sec. \ref{sec42}, unrolling the state $\bz_1$ with $k=1$ and $\lambda=1$ results in the largest robustness drop in the early state $\bz_3$.}
    \label{fig:ablation}
    \vspace{-0.4cm}
\end{figure*}

In this section, we study the effect of the white-box attacks with gradients estimated at different intermediate states in the forward solver of the large-sized DEQs. 

We first inspect the attacks with intermediate gradients acquired from each adjoint state. Fig. \ref{fig:ablation}-(a) plots the robustness of the early-state and final-state defense in both the unrolling-trained and the exact-trained DEQs. For the exact-trained DEQ, due to its violated fixed-point structure, $\bu_8$ results in the strongest attack for both the early-state and the final-state defenses. For the unrolling-trained DEQ, the estimated gradients at the consecutive adjoint states $\{\bu_n\}$ form increasingly stronger attacks on the robustness of the final state. On the robustness at the early state ($\bz_3$), the state $\bu_4$ gives rise to the strongest attack, which coincides with Eq. (\ref{adjoint-residual-v}) and Eq. (\ref{adjoint-update}) that $\bu_{n+1}$ directly depends on $\bz_n$. 

\begin{wrapfigure}{R}{0.4\textwidth}
    \vspace{-10pt}
    \centering
    \includegraphics[width=0.4\textwidth]{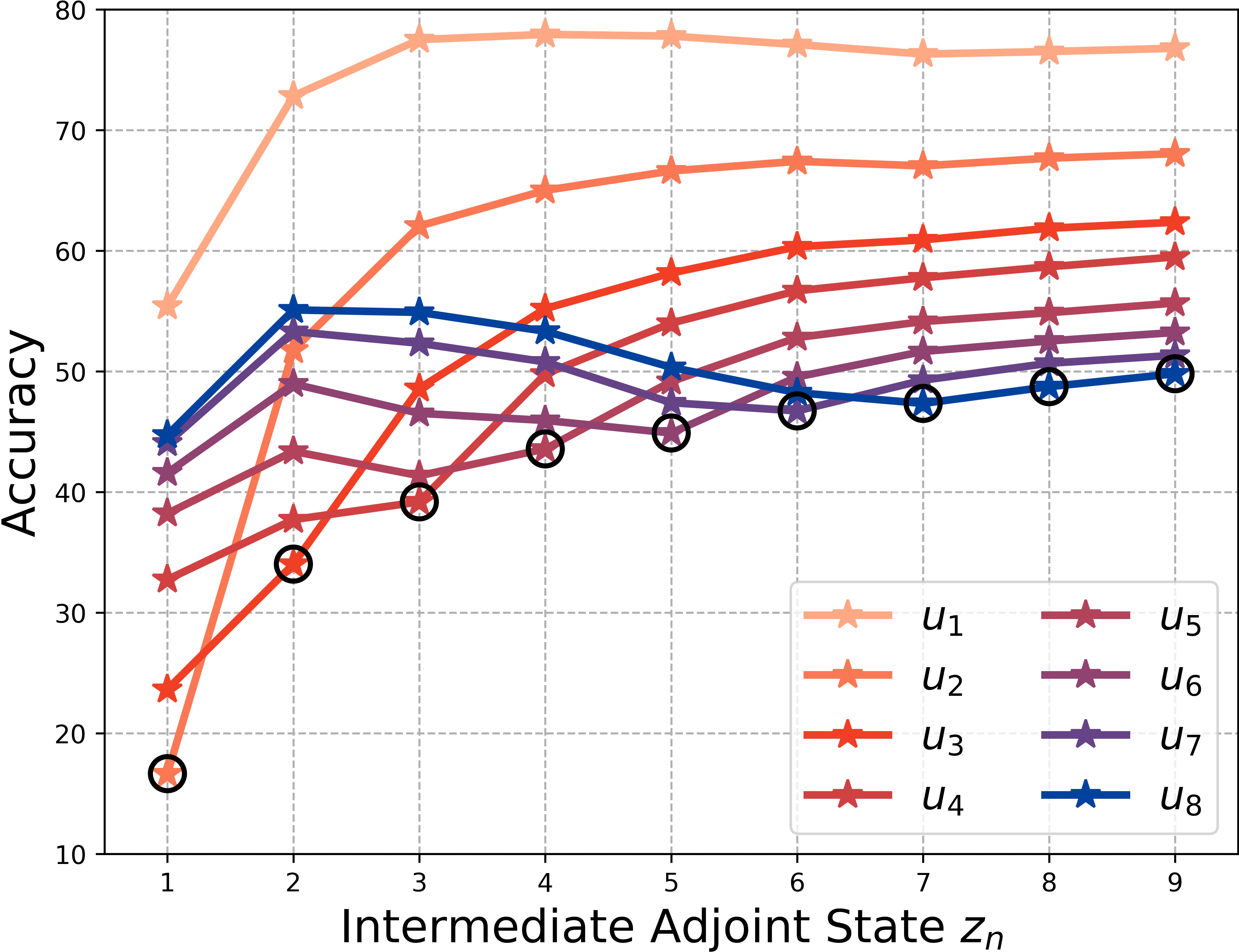}
    \caption{Alignment between the simultaneous adjoint process and the forward pass in the unrolling-trained DEQ.}\label{fig:simultaneous-adjoint-alignment}
    \vspace{-12pt}
\end{wrapfigure}
We further explore whether the simultaneous adjoint process are aligned with the forward pass. We use each intermediate adjoint state $\bu_n$ as the gradient surrogate in the PGD-10 attack for the unrolling-trained DEQ. Fig. \ref{fig:simultaneous-adjoint-alignment} shows the robustness performance of all the intermediate states $\bz_n$ in the forward pass of the unrolling-trained DEQ-Large. According to Fig. \ref{fig:simultaneous-adjoint-alignment}, it always follows that $\bu_{n+1}$ results in the largest robustness drop of $\bz_n$. As $\bu_{n+1}$ directly depends on $\bz_n$ in Eq. (\ref{adjoint-update}), this validates that the simultaneous adjoint process is aligned with the forward pass at each iteration in terms of adversarial robustness.

We also study the effect of unrolling different intermediate states $\{\bz_n\}$ for surrogate gradient estimation. Fig. \ref{fig:ablation}-(b)/(c) illustrates the robustness of the early state $\bz_3$ in the unrolling-trained DEQ under white-box attacks in different settings. It is shown that the number of unrolling steps for intermediate states like $\bz_1$ and $\bz_2$ should not be too much in order to obtain a powerful intermediate attack. The reason of this might be the inaccuracy of the unrolled intermediate gradient estimates.

The gradient estimated by unrolling $\bz_1$ leads to the most vigorous attack on the robustness of $\bz_3$. To understand the circumstance, we note that the unrolling-based intermediate gradient reflects only the feedback from the loss function at the unrolled state in Eq. (\ref{unroll-intermediate-gradient}) and Eq. (\ref{eq:unroll-intermediate-gradient-with-k}). As a result, the estimated gradient may be misaligned with the unrolled state: gradients by unrolling $\bz_3$ compose weak attack in terms of the robustness of $\bz_3$. It is inferred that the perturbation from the unrolled intermediate gradients must still be propagated in the forward pass to induce enough threatening distortion. This explains the delay that the unrolled gradient at $\bz_1$ affects the robustness of $\bz_3$. More ablation studies on the unrolled intermediates can be found in Appendix \ref{app:ablation-for-unrolled-intermediates}.

\subsection{Performance of the proposed attacks on vanilla DEQ models}\label{appendix-vanilla-deq} 

\begin{table}[t]
    \centering
    \caption{The performance (\%) of the standardly-trained DEQ-Large under ready-made PGD-10.}
    \resizebox{0.75\textwidth}{!}{\begin{tabular}{r|rrrrrrrr}
        \toprule
        State & $\bz_1$ & $\bz_2$ & $\bz_3$ & $\bz_4$ & $\bz_5$ & $\bz_6$ & $\bz_7$ & $\bz_8$ \\
        \midrule
        Clean Acc. & 38.81 & 82.62 & 89.63 & 91.77 & 92.08 & 92.29 & 92.39 & 92.53 \\
        Robust Acc. & 2.00 & 0.00 & 0.00 & 0.00 & 0.00 & 0.00 & 0.00 & 0.00 \\
        \bottomrule
    \end{tabular}}
    \label{tab:vanilla-ready-made-pgd10}
\end{table}

\begin{table*}[t]
    \caption{ Performance (\%) of the standardly-trained DEQ-Large \cite{jacobian-deq} with all proposed adaptive attacks and defense strategies under the PGD attack. The notations for the rows and the columns are similar with those in Tables 2 and 3. Under the (\underline{underlined}) strongest attacks, the ensemble defense still achieves the best robustness performance (\textbf{in bold}).}
    \vspace{-0.35cm}
    \label{tab:vanilla-trained-deq}
    \begin{center}
    \resizebox{0.85\textwidth}{!}{\begin{tabular}{cccccccc}
        \toprule
        \multirow{2}{*}{Defense} & \multirow{2}{*}{Clean} & \multicolumn{3}{c}{Simultaneous Adjoint} & \multicolumn{3}{c}{Unrolled Intermediates} \\
        \cmidrule{3-8}
        ~ & ~ & Final & Intermediate & Ensemble & Final & Intermediate & Ensemble \\
        \midrule
        Final    & 92.53 & 8.90 & 11.45 & 3.69 & \underline{0.00} & \underline{0.00} & \underline{0.00} \\
        Early    & 38.81 & 6.08 & \;4.54 & 3.42 & 2.00 & 2.94 & \underline{\textbf{1.31}} \\
        Ensemble & 87.31 & 9.12 & \;6.39 & 3.48 & \underline{0.00} & \underline{0.00} & \underline{0.00} \\
        \bottomrule
    \end{tabular}}
    \end{center}
    \vspace{-0.5cm}
\end{table*}

In this section, we evaluate the performance of the proposed attacks on the DEQ models without adversarial training. We train a DEQ-Large on CIFAR-10 with standard training following the recipe in \cite{jacobian-deq}, and use the ready-made PGD-10 to attack the model. The clean accuracy of each state $\bz_n$, as well as its robust accuracy is shown in Table \ref{tab:vanilla-ready-made-pgd10}. Different from the robustness accumulation effect in the adversarially-trained DEQs (shown in Sec. \ref{sec:compare-exact-unrolling-under-white-attacks}, Fig. \ref{fig:figure1-real}-(a), and Appendix \ref{app:rob-accum-xl}), the ready-made PGD-10 already has a dramatic effect in attacking all the states in the standardly-trained DEQ. 

We proceed to apply all the proposed attacks and defense strategies. Following Sec. \ref{section:white-box-defenses}, we determine the optimal timing for early exiting the standardly-trained DEQ as state $\bz_1$. Shown in Table \ref{tab:vanilla-trained-deq}, it can be seen that all the proposed attacks can defeat the DEQ by standard training. As the white-box robustness of DEQs is assessed by the strongest defense under all attacks (minimum over all columns in a row, then maximum over the minimum of the rows), the white-box robustness of the vanilla DEQ is 1.31\% with a 38.81\% clean accuracy using the early-state defense. When using the final-state and the ensemble-state defense, the robustness is 0\%. These results validate that all the proposed attacks are reliably strong as they all defeat the DEQ models without adversarial training. 

\section{Conclusion} \label{section-conclusion}

We study the adversarial robustness of general DEQs, using the exact gradient and the unrolling-based phantom gradient in adversarial training for DEQs, respectively. We observe the gradient obfuscation issues in DEQs under ready-made attacks. Based on the misalignment between the forward and backward tracks, we leverage intermediate states in the forward pass to construct white-box attacks and defense strategies and benchmark the white-box robustness performance of DEQs. 

While we have performed a serious comparison of white-box robustness between DEQs and deep networks, it can be seen that the performance of DEQs is on par with that of deep networks. Our empirical observations indicate that we should explore more advanced AT mechanisms for DEQs, in order to exploit their local attractor structures. A potential way is to explicitly encourage closed-loop control during training, similar to the mechanism introduced in \cite{chen2021towards}. To this end, the gradient estimation method proposed in this paper would be one of the critical ingredients for solving the misalignment between the forward/backward pass of DEQs.

\section*{Acknowledgements}

This work was supported by the National Natural Science Foundation of China (No.61925601) and Beijing Academy of Artificial Intelligence (BAAI). We appreciate all of the anonymous reviewers for their comments and suggestions on this work.

\clearpage
\bibliography{neurips22_yzh}
\bibliographystyle{plain}

\clearpage

\section*{Checklist}

\begin{enumerate}

\item For all authors...
\begin{enumerate}
  \item Do the main claims made in the abstract and introduction accurately reflect the paper's contributions and scope?
    \answerYes{}
  \item Did you describe the limitations of your work?
    \answerYes{We describe the future work in the Conclusion section. We attach analysis and discuss about the limitations of our work in Appendices.}
  \item Did you discuss any potential negative societal impacts of your work?
    \answerYes{We discuss them in the Broad Impact section.}
  \item Have you read the ethics review guidelines and ensured that your paper conforms to them?
    \answerYes{}
\end{enumerate}

\item If you are including theoretical results...
\begin{enumerate}
  \item Did you state the full set of assumptions of all theoretical results?
    \answerYes{}
        \item Did you include complete proofs of all theoretical results?
    \answerYes{}
\end{enumerate}

\item If you ran experiments...
\begin{enumerate}
  \item Did you include the code, data, and instructions needed to reproduce the main experimental results (either in the supplemental material or as a URL)?
    \answerYes{}
  \item Did you specify all the training details (e.g., data splits, hyperparameters, how they were chosen)?
    \answerYes{}
        \item Did you report error bars (e.g., with respect to the random seed after running experiments multiple times)?
    \answerNo{}
        \item Did you include the total amount of compute and the type of resources used (e.g., type of GPUs, internal cluster, or cloud provider)?
    \answerYes{}
\end{enumerate}

\item If you are using existing assets (e.g., code, data, models) or curating/releasing new assets...
\begin{enumerate}
  \item If your work uses existing assets, did you cite the creators?
    \answerYes{}
  \item Did you mention the license of the assets?
    \answerNA{}
  \item Did you include any new assets either in the supplemental material or as a URL?
    \answerNA{}
  \item Did you discuss whether and how consent was obtained from people whose data you're using/curating?
    \answerNA{}
  \item Did you discuss whether the data you are using/curating contains personally identifiable information or offensive content?
    \answerNA{}
\end{enumerate}

\item If you used crowdsourcing or conducted research with human subjects...
\begin{enumerate}
  \item Did you include the full text of instructions given to participants and screenshots, if applicable?
    \answerNA{}
  \item Did you describe any potential participant risks, with links to Institutional Review Board (IRB) approvals, if applicable?
    \answerNA{}
  \item Did you include the estimated hourly wage paid to participants and the total amount spent on participant compensation?
    \answerNA{}
\end{enumerate}

\end{enumerate}



\newpage
\appendix
\onecolumn
\section{Experimental settings} \label{app:experimental-settings}

 We train DEQs on CIFAR-10 with the PGD-AT framework. The detailed experimental settings for both the unrolling-trained and the exact-trained DEQs are shown in Table \ref{tab:hyperparams}. The best checkpoint is selected with the top performance on the development set under the ready-made PGD-10 attack. The gradient used in the attack for checkpoint selection is the same as the gradient used for training. The settings on the basics of DEQ training are largely followed from \cite{jacobian-deq}. We leave the integration of the estimated intermediate gradients for adversary generation into AT for DEQs as future work. We use NVIDIA-3090 GPUs for all of our experiments.

\begin{table*}[ht]
    \caption{Detailed hyperparameter settings.}
    \label{tab:hyperparams}    
    \centering
    \resizebox{\textwidth}{!}{\begin{tabular}{llcc}
        \toprule
        Category & Settings & DEQ-Large & DEQ-XL \\
        \midrule
        \multirow{7}{*}{Architecture} & Input Image Size & \multicolumn{2}{c}{32 $\times$ 32} \\
        ~ & Number of Scales & \multicolumn{2}{c}{4} \\
        ~ & \# of Head Channels for Each Scale & [14, 28, 56, 112] & [20, 40, 80, 160] \\
        ~ & \# of Channels for Each Scale & [32, 64, 128, 256] & [72, 144, 288, 576] \\
        ~ & Channel Size of Final Layer & 1,680 & 1,800 \\
        ~ & Activation Function & \multicolumn{2}{c}{ReLU} \\               
        ~ & \# of Parameters & 10M & 48M \\
        \midrule
        \multirow{4}{*}{DEQ Solver} & \# of Forward Solver Iterations & \multicolumn{2}{c}{8} \\
        ~ & \# of Backward Solver Iterations & \multicolumn{2}{c}{7} \\
        ~ & Algo. for Forward Solvers & \multicolumn{2}{c}{Broyden's Method} \\
        ~ & Algo. for Backward Solvers & \multicolumn{2}{c}{Broyden's Method} \\
        \midrule
        \multirow{7}{*}{Optimization} & Optimizer & \multicolumn{2}{c}{Adam} \\
        ~ & Learning Rate Schedule & \multicolumn{2}{c}{cosine decay} \\
        ~ & Decay Factor & \multicolumn{2}{c}{0.1} \\
        ~ & Epochs for Decay & \multicolumn{2}{c}{[30, 60, 90]} \\
        ~ & Initial Learning Rate & \multicolumn{2}{c}{0.001} \\
        ~ & Nesterov Momentum & \multicolumn{2}{c}{0.98} \\
        ~ & Weight Decay & \multicolumn{2}{c}{-} \\
        \midrule
        \multirow{12}{*}{Adv. Training} & Batch Size & \multicolumn{2}{c}{96} \\
        ~ & Training Epochs & \multicolumn{2}{c}{150} \\
        ~ & Pretraining Steps & \multicolumn{2}{c}{16,000} \\
        \cmidrule{2-4}
        ~ & Weight of JR During Pretraining & \multicolumn{2}{c}{-} \\
        ~ & Weight of JR During DEQ Training & \multicolumn{2}{c}{0.4} \\
        ~ & Stop Epoch for JR & \multicolumn{2}{c}{90} \\ 
        \cmidrule{2-4}
        ~ & Unrolling-Trained: Steps $k$  & \multicolumn{2}{c}{5} \\
        ~ & Unrolling-Trained: Damping Factor $\lambda$ & \multicolumn{2}{c}{0.5} \\ 
        \cmidrule{2-4}
        ~ & AT for Pretraining & \multicolumn{2}{c}{No} \\        
        ~ & Attack in AT for DEQ Training & \multicolumn{2}{c}{ready-made PGD-10} \\
        ~ & Grad. for Adv. Generation in AT & \multicolumn{2}{c}{same with the gradient used in training} \\
        ~ & Label Smoothing & \multicolumn{2}{c}{-} \\
        \bottomrule
    \end{tabular}}
\end{table*}




\newpage

\section{On the convergence of simultaneous adjoint } \label{app:theorem-proof}

We start with the assumption on nonsingularity of the Jacobian inverse of $g_\theta(\bz_n;x) = f_\theta(\bz_n;x) - \bz_n$:
\begin{assumption}
    For $\forall \bz \in \mathbb{R}^{d}$ and $\forall \bx \in \mathbb{R}^{l}$, we assume the nonsingularity of Jacobian inverse of $g_\theta(\bz;x) = f_\theta(\bz;x) - \bz$. Namely,
    \begin{equation}
        \left(\frac{\partial g_\theta(\bz;x)}{\partial \bz} \right)^{-1} = \left(\frac{\partial f_\theta(\bz; \bx)}{\partial \bz} - I\right)^{-1}
    \end{equation}
    exists.
\label{the-assumption-0}
\end{assumption}
We also make assumptions on the precision of $B_n$, which is the approximation of the Jacobian inverse of $g_\theta(\bz_n;x)$.
\begin{assumption}
    Define
    \begin{equation}
        \Bar{B}_n \left(\frac{\partial g_\theta(\bz_n;x)}{\bz_n} \right) = I - \boldsymbol{\epsilon}_n(\bx).
    \end{equation}
    We assume that  $\|\boldsymbol{\epsilon}_n(\bx)\| \le 1 - \epsilon < 1$ with $\epsilon > 0$ for all $\bx$.
\label{the-assumption}
\end{assumption}
With these two assumptions, with $0 < \beta < 1$, we have:
\begin{align}
    & \; \left \| \bu_{n+1} - \bu^* \right \| \\
    = & \; \left\| \bu_n - \beta \Bar{B}_n \bv_n - \bu^* \right \| \\
    = & \;  \left \| \bu_n - \beta \Bar{B}_n \left(\left(\frac{\partial f_\theta(\bz_n; \bx)}{\partial \bz}\right) \bu_n + \frac{\partial L(\bz_n, y)}{\partial \bz} - \bu_n\right) - \bu^* \right \| \\
    = & \;  \left\| \bu_n - \beta \Bar{B}_n \left(\frac{\partial g_\theta(\bz_n;x)}{\partial \bz} \right) \bu_n - \beta \Bar{B}_n \frac{\partial L(\bz_n, y)}{\partial \bz} - \bu^* \right \| \\
    = & \;  \left\| \bu_n - \beta \left( I - \boldsymbol{\epsilon}_n(\bx) \right) \bu_n - \beta \Bar{B}_n \frac{\partial L(\bz_n, y)}{\partial \bz} - \bu^* \right \| \\
    = & \; \left \| \left((1 - \beta) I + \beta \boldsymbol{\epsilon}_n(\bx) \right) \bu_n + \beta \left( I - \boldsymbol{\epsilon}_n(\bx) \right) \left( I - \frac{\partial f_\theta(\bz_n; \bx)}{\partial \bz} \right)^{-1} \frac{\partial L(\bz_n, y)}{\partial \bz} - \bu^* \right \| \\ 
    = & \; \left \| \left((1 - \beta) I + \beta \boldsymbol{\epsilon}_n(\bx) \right) (\bu_n - \bu^*) + \beta \left( I - \boldsymbol{\epsilon}_n(\bx) \right) \left( \left( I - \frac{\partial f_\theta(\bz_n; \bx)}{\partial \bz} \right)^{-1} \frac{\partial L(\bz_n, y)}{\partial \bz} - \bu^* \right) \right \| \\ 
    \le & \; \left \| (1 - \beta) I + \beta \boldsymbol{\epsilon}_n(\bx) \right \| \left \| \bu_n - \bu^* \right \| + \beta \left\| I - \boldsymbol{\epsilon}_n(\bx) \right\| \left\| \left( I - \frac{\partial f_\theta(\bz_n; \bx)}{\partial \bz} \right)^{-1} \frac{\partial L(\bz_n, y)}{\partial \bz} - \bu^* \right \|. \label{two-term-proof} 
\end{align}

Define
\begin{equation}
    \bw_n = \left( I - \frac{\partial f_\theta(\bz_n; \bx)}{\partial \bz} \right)^{-1} \frac{\partial L(\bz_n, y)}{\partial \bz}.
\end{equation}
According to the definition of $\bu^*$ in Eq. (\ref{exact-gradient}), it follows that $\bw_n \to \bu^*$ when $\bz_n \to \bz^*$. $\{\bw_n\}$ has the highest convergence rate to $\bu^*$ as it is the most precised estimation of $\bu^*$ at step $n$.
According to Assumption \ref{the-assumption} and $0 < \beta < 1$, the first term in Eq. (\ref{two-term-proof}) follows
\begin{equation}
    \left \| (1 - \beta) I + \beta \boldsymbol{\epsilon}_n(\bx) \right \| \le (1-\beta) + \beta \| \boldsymbol{\epsilon}_n(\bx) \| \le 1-\beta\epsilon < 1 . 
\label{proof-first-term}
\end{equation}

Substituting Eq. (\ref{proof-first-term}) and the upper bound $\epsilon_0$ into Eq. (\ref{two-term-proof}), we have
\begin{equation}
    \left \| \bu_{n+1} - \bu^* \right \| \le (1-\beta\epsilon) \left \| \bu_{n} - \bu^* \right \| + \| \bw_n - \bu^* \|,
\end{equation}
thus
\begin{equation}
    \frac{\left \| \bu_{n+1} - \bu^* \right \|}{\left \| \bu_{n} - \bu^* \right \|} \le 1-\beta\epsilon + \frac{\| \bw_n - \bu^* \|}{\left \| \bu_{n} - \bu^* \right \|}.
\label{final-inequality}
\end{equation}
We omit the second term in the right-hand side of (\ref{final-inequality}) under the mild assumption of the strictly higher convergence rate of $\{\bw_n\}$. As a result, when $n>N$ with $N$ sufficiently large, it follows that $\{\bu_n\}$ converges to $\bu^*$ as $\left \| \bu_{n+1} - \bu^* \right \| < \left \| \bu_{n} - \bu^* \right \|$. However, we do \textit{not} require the convergence of $\bu_n$ as we only use them in Eq. (\ref{sm-adjoint-gradient}) as gradient \textit{estimations} to construct \textit{intermediate attacks}. In practice, we tune $\beta$ to facilitate the strongest attacks. For the (8/7) unrolling-trained DEQ-Large/XL, we set $\beta=0.5$; for the (50/50) exact-trained DEQ-Large, we set $\beta=0.05$.

\section{The robustness accumulation effect} \label{app:rob-accum-xl}

\begin{figure*}[ht]
    \centering
    \includegraphics[width=\textwidth]{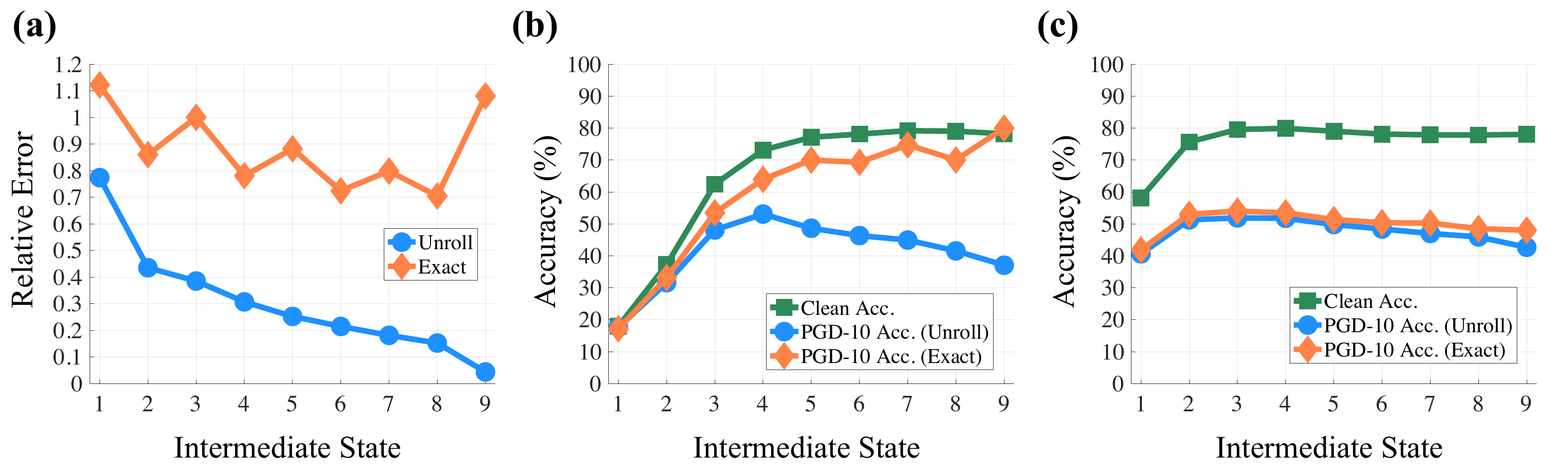}
    \caption{ The performance of all intermediate states $\bz_n$ in the exact-trained and the unrolling-trained DEQ-Large with 8 forward and 7 backward iterations. (a) Relative error at each intermediate state $\bz_n$. The fixed-point structure in the exact-trained DEQ is violated as all relative errors are higher than $1.0$. (b) and (c): Robustness evaluated at different intermediate states in (b) the exact-trained and (c) the unrolling-trained DEQs. The unrolling-based phantom gradient forms stronger adversaries in the ready-made PGD-10 attack. The intermediate state $\bz_3$ exhibits the strongest robustness in the unrolling-trained DEQ, while $\bz_4$ exhibits the strongest robustness in the exact-trained DEQ.}
    \label{fig:train-exact-train-unroll-large}
\end{figure*}

\begin{figure*}[ht]
    \centering
    \includegraphics[width=\textwidth]{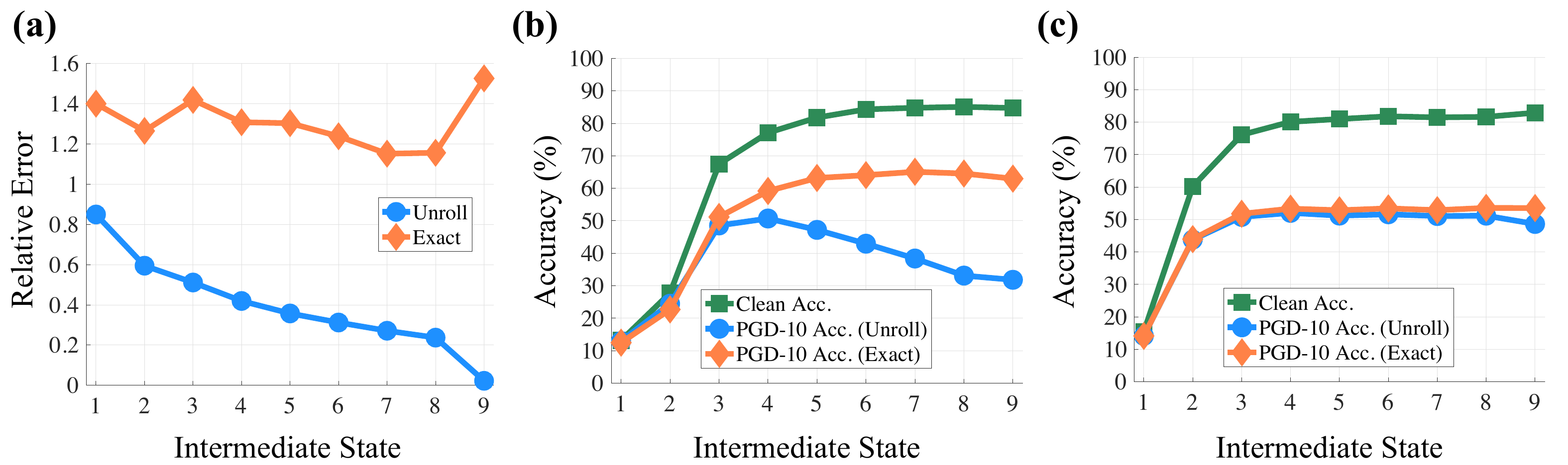}
    \caption{ The performance of all intermediate states $\bz_n$ in the exact-trained and the unrolling-trained DEQ-XL with 8 forward and 7 backward iterations. (a) Relative error at each intermediate state $\bz_n$. The fixed-point structure in the exact-trained DEQ is violated as all relative errors are higher than $1.0$. (b) and (c): Robustness evaluated at different intermediate states in (b) the exact-trained and (c) the unrolling-trained DEQs. The unrolling-based phantom gradient forms stronger adversaries in the ready-made PGD-10 attack. For both DEQs, the intermediate state $\bz_4$ exhibits stronger robustness than other states.}
    \label{fig:train-exact-train-unroll-xl}
\end{figure*}

Figures \ref{fig:train-exact-train-unroll-large} and \ref{fig:train-exact-train-unroll-xl} the robustness accumulation effect in the DEQs with 8 forward and 7 backward iterations. For both the exact-trained and the unrolling-trained DEQs, the robustness accumulation effect exists because the black-box solvers have obfuscated the gradients used in the ready-made attacks. It can also be seen that the fixed-point structures are broken in the exact-trained DEQ-Large and DEQ-XL under this setting. Figure \ref{fig:figure1-real} show that the exact-trained DEQs with more iterations in the solvers can retain the fixed-point structure (more details in Appendix \ref{appendix-f-2}). 

\clearpage
\section{Comparison with the adjoint Broyden method} \label{app:adjoint-broyden}

Similar to our simultaneous adjoint design, concurrent work \cite{SHINE} propose the adjoint Broyden method to share the approximated Jacobian inverse $B_n$ into the backward pass. Their motivation, however, is to accelerate DEQ training while we integrate the simultaneous adjoint into ready-made attacks to facilitate white-box robustness evaluation. In this section, we compare the adjoint Broyden method with our simultaneous adjoint in terms of the effect of intermediate/ensemble attacks.

\begin{table*}[h]
    \caption{Comparison between the proposed simultaneous adjoint and the adjoint Broyden method.}
    \label{tab:sa-ab}
    \begin{center}
    \begin{tabular}{cccccc}
        \toprule
        \multirow{2}{*}{Arch.} & \multirow{2}{*}{Defense} & \multirow{2}{*}{Clean} & \multicolumn{3}{c}{PGD : Simultaneous Adjoint / Adjoint Broyden} \\
        \cmidrule{4-6}
        ~ & ~ & ~ & Final & Intermediate & Ensemble \\
        \midrule
        \multirow{3}{*}{DEQ-Large} & Final & 78.03 & \textbf{49.81}/52.34 & 59.49/\textbf{58.22} & \textbf{54.91}/58.06 \\
        ~ & Early & 79.57 & 54.90/\textbf{45.29} & \textbf{39.19}/43.48 & \textbf{42.76}/45.47 \\
        ~ & Ensemble & 79.67 & 51.52/\textbf{49.24} & 52.43/\textbf{49.24} & \textbf{49.47}/53.04  \\
        \midrule
        \multirow{3}{*}{DEQ-XL} & Final & 82.92 & \textbf{55.80}/65.00 & \textbf{55.80}/65.00 & \textbf{67.30}/71.20 \\
        ~ & Early & 80.12 & \textbf{51.40}/59.26 & \textbf{51.40}/59.26 & \textbf{60.78}/65.67 \\
        ~ & Ensemble & 81.17 & \textbf{52.87}/60.22 & \textbf{52.87}/60.22 & \textbf{61.40}/66.69 \\
        \bottomrule
    \end{tabular}
    \end{center}
\end{table*}

Table \ref{tab:sa-ab} shows the comparison between our simultaneous adjoint and the adjoint Broyden method with PGD as the attack. The proposed simultaneous adjoint results in stronger white-box attacks in general. The adjoint Broyden method also yields better performance in some cases. We further compare among the proposed simultaneous adjoint, the adjoint Broyden method, and the unrolled intermediates in Table \ref{tab:sa-ab-ui}. 

\begin{table*}[h]
    \caption{Comparison among the proposed simultaneous adjoint, the adjoint Broyden method, and the (proposed) unrolled intermediates with PGD.}
    \label{tab:sa-ab-ui}
    \begin{center}
    \resizebox{\textwidth}{!}{\begin{tabular}{cccccc}
        \toprule
        \multirow{2}{*}{Arch.} & \multirow{2}{*}{Defense} & \multirow{2}{*}{Clean} & \multicolumn{3}{c}{PGD : Simultaneous Adjoint / Adjoint Broyden / Unrolling Intermediates} \\
        \cmidrule{4-6}
        ~ & ~ & ~ & Final & Intermediate & Ensemble \\
        \midrule
        \multirow{3}{*}{DEQ-Large} & Final & 78.03 & 49.81/52.34/\textbf{42.67} & 59.49/\textbf{58.22}/62.24 & 54.91/58.06/\textbf{51.52} \\
        ~ & Early & 79.57 & 54.90/\textbf{45.29}/51.90 & 39.19/43.48/\textbf{29.38} & 42.76/45.47/\textbf{34.20} \\
        ~ & Ensemble & 79.67 & 51.52/49.24/\textbf{49.02} & 52.43/\textbf{49.24}/55.10 & 49.47/53.04/\textbf{47.12}  \\
        \midrule
        \multirow{3}{*}{DEQ-XL} & Final & 82.92 & 55.80/65.00/\textbf{48.58} & \textbf{55.80}/65.00/65.94 & 67.30/71.20/\textbf{58.23} \\
        ~ & Early & 80.12 & \textbf{51.40}/59.26/52.08 & \textbf{51.40}/59.26/55.70 & 60.78/65.67/\textbf{48.88} \\
        ~ & Ensemble & 81.17 & 52.87/60.22/\textbf{51.70} & \textbf{52.87}/60.22/59.71 & 61.40/66.69/\textbf{56.45} \\
        \bottomrule
    \end{tabular}}
    \end{center}
\end{table*}

Table \ref{tab:sa-ab-ui} shows that the unrolled intermediates still have the dominant effect in facilitating white-box attacks in general. In the future work, we will integrate our simultaneous adjoint/the adjoint Broyden method into the adversarial training process of DEQs and validate their white-box robustness.

\clearpage
\section{Ablation studies}
\subsection{Settings of the unrolled intermediates} \label{app:ablation-for-unrolled-intermediates}

\begin{figure*}[ht]
    \centering
    \subfigure[Exact-trained, defense with the final state]{
    \includegraphics[width=0.45\textwidth]{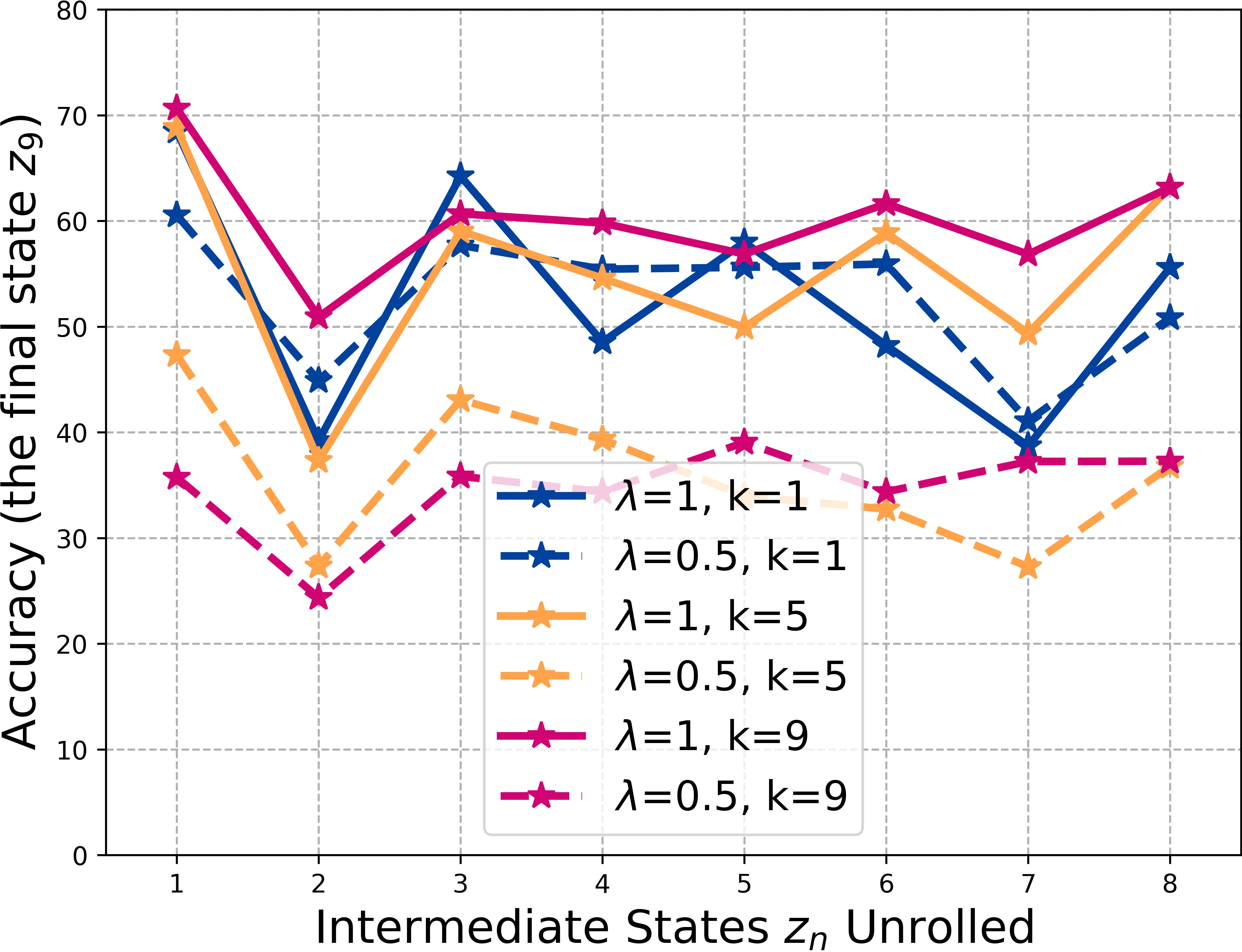}
    }
    \quad
    \subfigure[Exact-trained, defense with the early state]{
    \includegraphics[width=0.45\textwidth]{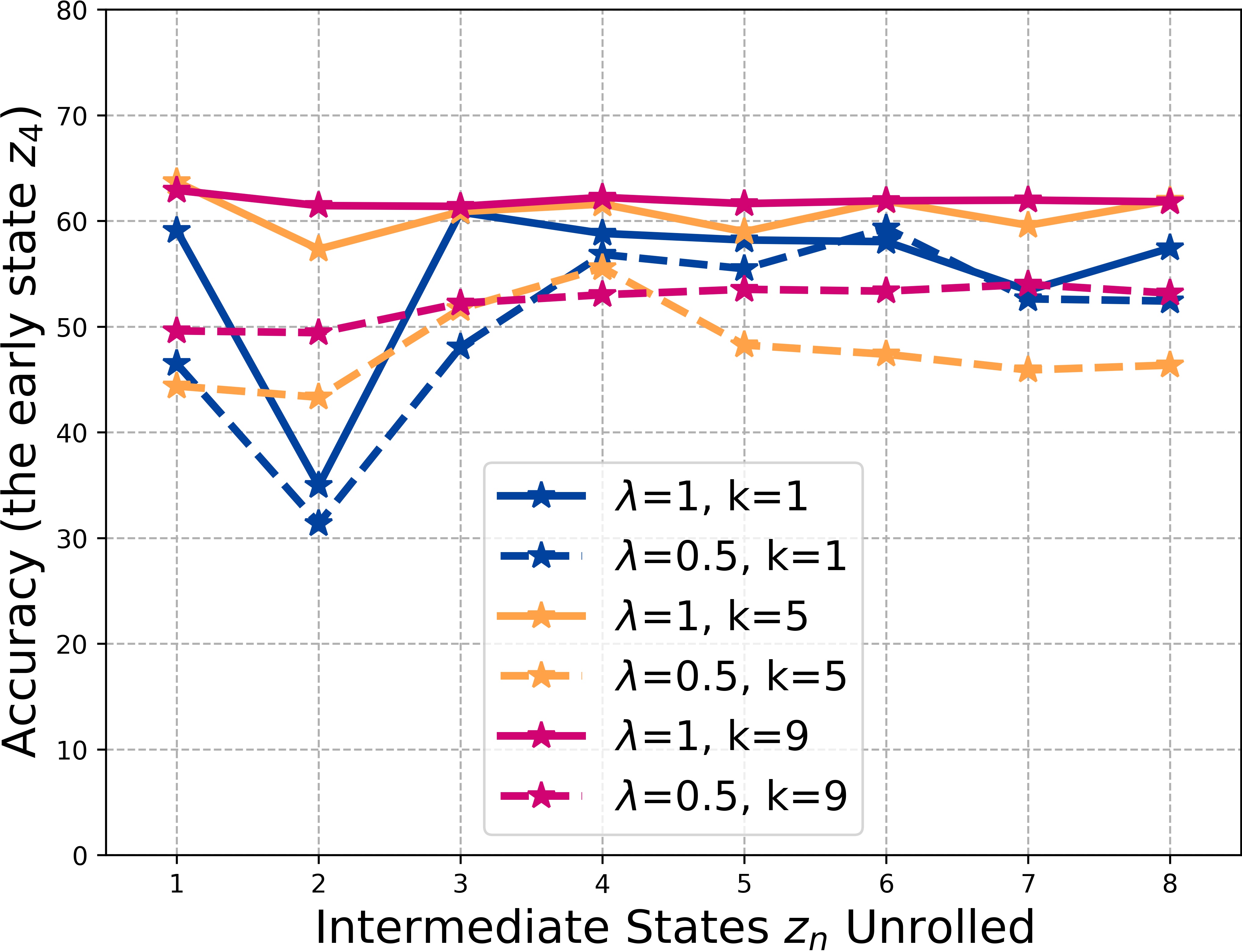}
    }
    \caption{Ablation on unrolling different intermediate states with different steps in exact-trained DEQ-Large.}
    \label{fig:ablation-unroll-intermediate-exact-trained}
\end{figure*}

\begin{figure*}[ht]
    \centering
    \subfigure[Unrolling-trained, defense with the final state]{ 
    \includegraphics[width=0.45\textwidth]{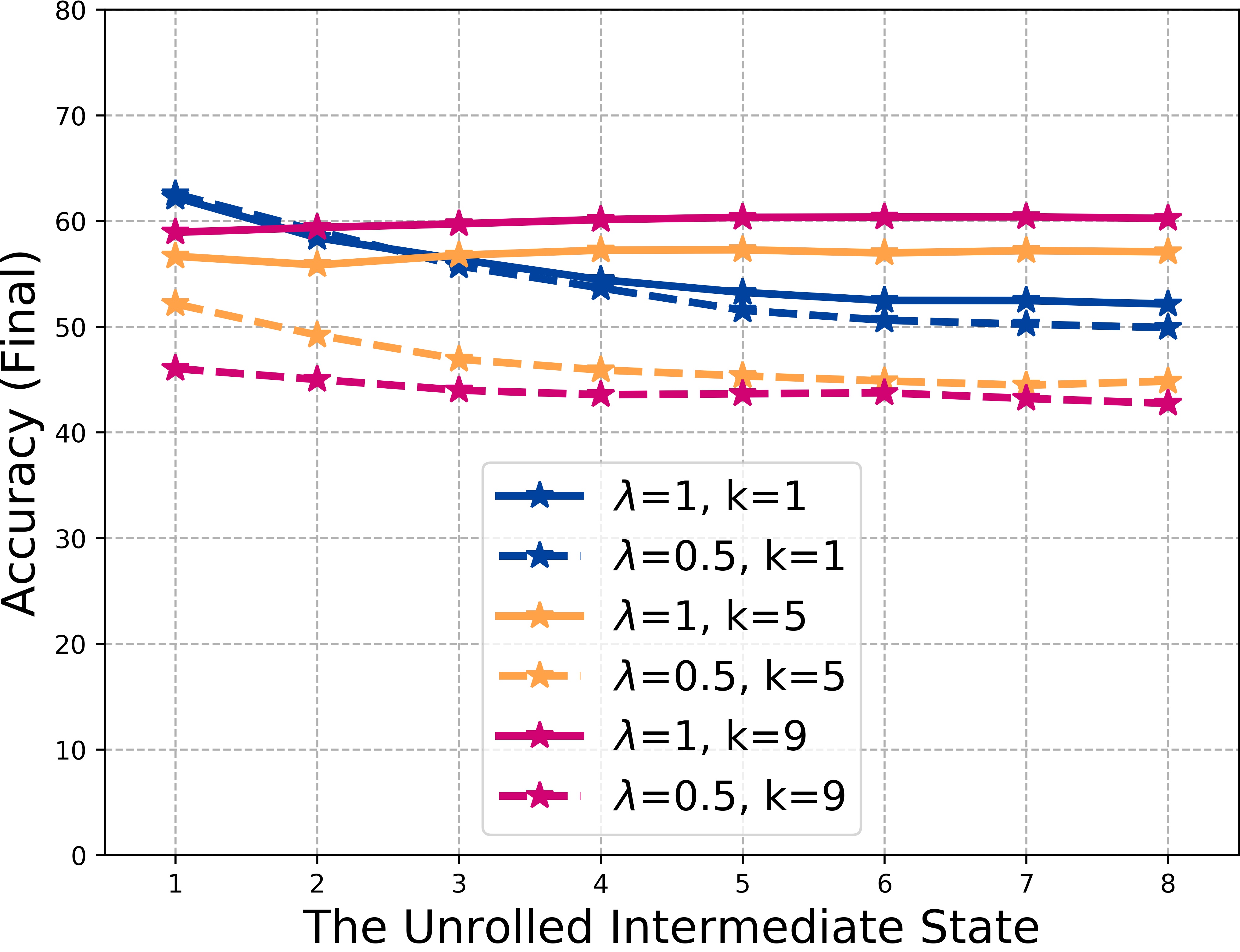}
    }
    \quad
    \subfigure[Unrolling-trained, defense with the final state]{ 
    \includegraphics[width=0.45\textwidth]{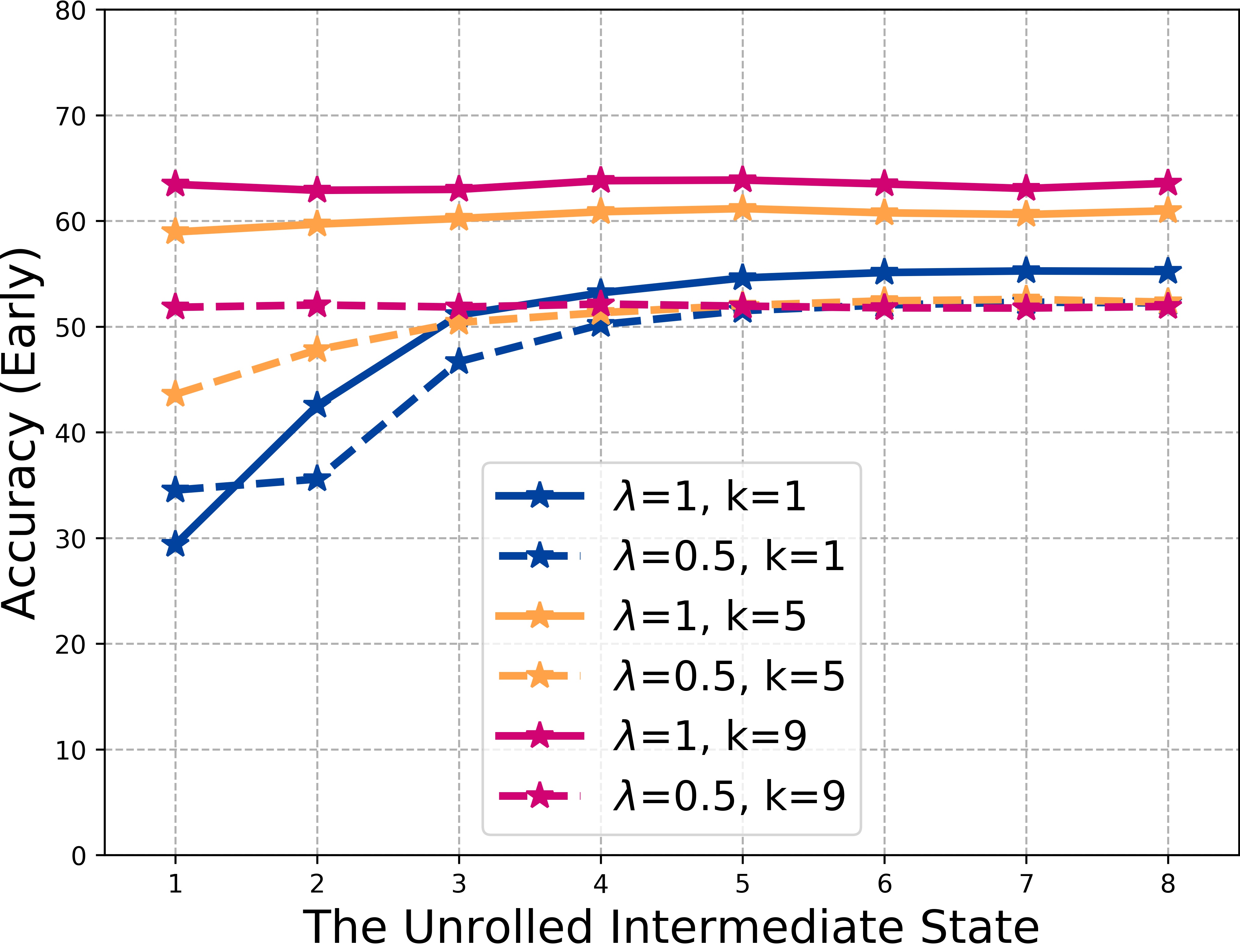}
    }
    \caption{Ablation on unrolling different intermediate states with different steps in unrolling-trained DEQ-Large.}
    \label{fig:ablation-unroll-intermediate-unrolling-trained}
\end{figure*}

\subsection{The backward solver iteration threshold in attacks with exact gradients}\label{appendix-f-1}

The exact gradient in Eq. (\ref{exact-gradient}) is solved by an independent fixed-point iteration process solely based on $\bz^*$. While this iteration process is not aligned with the forward pass, one might still question whether the intermediate states in the backward solver can lead to strong attacks.

\begin{figure*}[ht]
    \centering
    \includegraphics[width=0.6\textwidth]{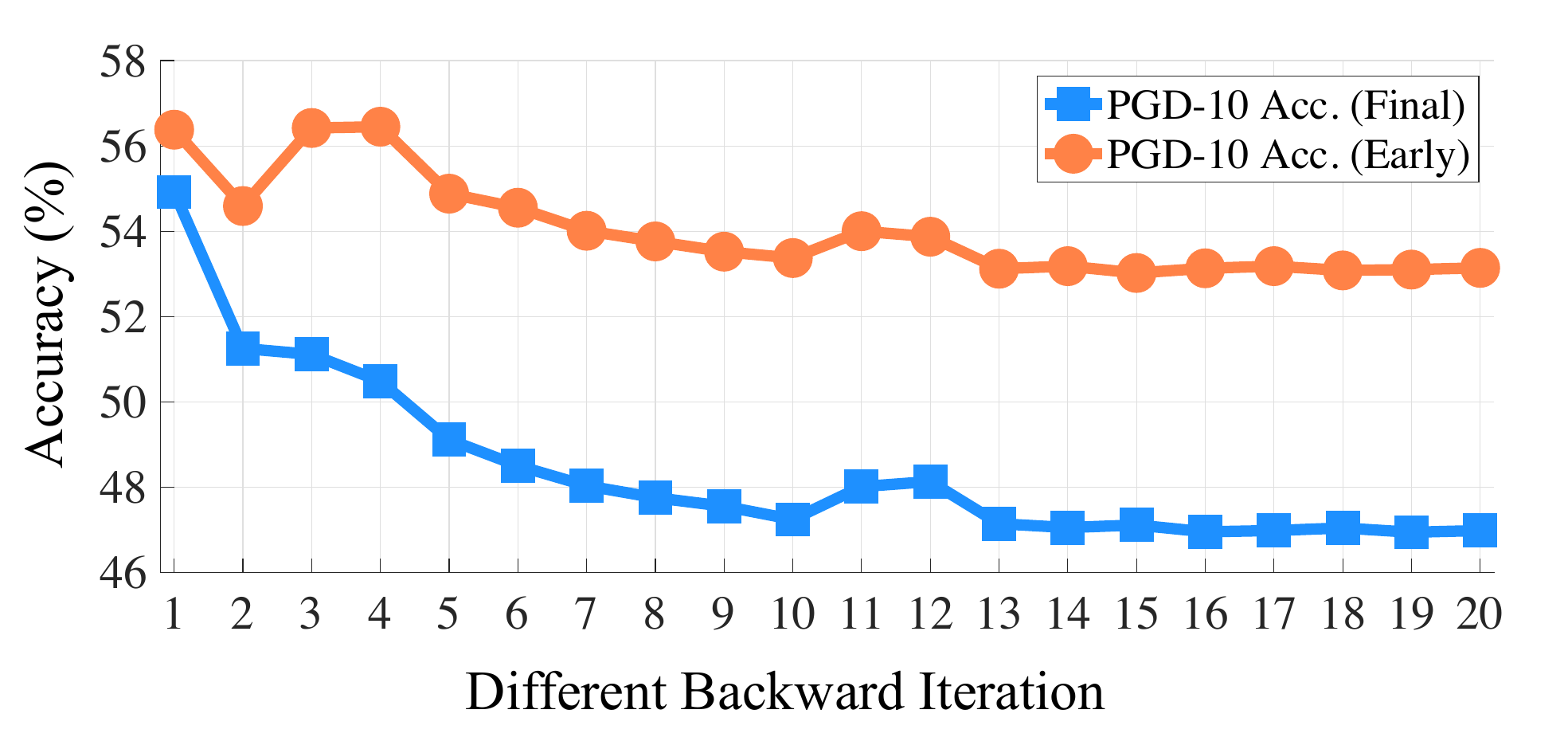}
    \caption{Robustness of the final state as well as the early state in the unrolling-trained DEQ-Large under PGD-10. The gradients in the attacks are returned from the backward solver with different iterations.}
    \label{fig:diff-backiter}
\end{figure*}

We address the question by experimenting with the unrolling-trained DEQ-Large. We use the exact gradient solved by a backward solver for adversary generation in PGD-10. While the default iteration threshold is $7$, we investigate the capability of all the $7$ intermediate states in constructing adversaries. We also increase the number of the iterations up to $20$ to see whether the robustness is degraded. The results are shown in Fig. \ref{fig:diff-backiter}.

\looseness=-1 From Fig. \ref{fig:diff-backiter}, it can be seen that the gradients from less iterations result in less powerful attacks for both the final and the early states. With increased backward iterations, the robustness of both states is dropped by about $1$\%. However, as shown in Table \ref{tab:train-unroll-train-exact-robust-acc-pgd}, PGD-10 with the unrolling-based phantom gradient results in the robustness of $42.67$ for the final state and $51.90$ for the early state, both of which lower than those in Fig. \ref{fig:diff-backiter}. It is thus concluded that alternating the backward iterations in the exact gradient does not have significant effect on deteriorating the robustness of the unrolling-trained DEQ.

\subsection{Solver iteration thresholds in the exact-trained DEQs}\label{appendix-f-2}

We increase the iteration thresholds of the forward and the backward solvers in the exact-trained DEQs to stabilize adversarial training. We experiment with the exact-trained DEQ-Large and report the results in Table \ref{tab:exact-trained-iter-threshold} (which is also illustrated in Fig. \ref{fig:figure1-real}-(a)).

\begin{table*}[ht]
    \caption{Exact-trained DEQ-Large with larger iteration thresholds in the forward and the backward solvers during training. The ``PGD, Unroll" column, demonstrating the accuracy of the final and the top intermediate state under ready-made PGD attacks with the unrolling-based phantom gradient, is used to illustrate Fig. \ref{fig:figure1-real}-(a).}
    \label{tab:exact-trained-iter-threshold}
    \centering
    \resizebox{\textwidth}{!}{\begin{tabular}{clcrrrrc}
    \toprule
        Training & (ForIter,/BackIter) & Defense & Clean & PGD, Exact & PGD, Unroll & Speed (Samples/s) & Note \\
    \midrule
        \multirow{8}{*}{Exact}  & \multirow{2}{*}{$(8/7)$} & Final & 78.24 & 79.97 & 37.07 & \multirow{2}{*}{28.3} & \multirow{2}{*}{Grad. Obfus.} \\
        ~  & ~ & Early & 73.12 & 63.95 & 53.09 & ~ & ~\\
    \cmidrule{2-8}
        ~  & \multirow{2}{*}{$(18/20)$} & Final & 73.16 & 39.92 & 40.00 & \multirow{2}{*}{22.3} & \multirow{2}{*}{-}\\
        ~  & ~ & Early & 72.02 & 51.44 & 51.07 & ~ & ~  \\
    \cmidrule{2-8}
        ~  & \multirow{2}{*}{$(30/30)$} & Final & 81.17 & 41.51 & 37.12 & \multirow{2}{*}{7.6} & \multirow{2}{*}{-}\\
        ~  & ~ & Early & 87.72 & 65.22 & 62.74 & ~ & ~  \\     
    \cmidrule{2-8}
        ~  & \multirow{2}{*}{$(50/50)$} & Final & 73.51 & 37.29 & 36.70 & \multirow{2}{*}{5.3} & \multirow{2}{*}{-}\\
        ~  & ~ & Early & 86.98 & 74.04 & 73.93 & ~ & ~  \\        
    \midrule
        \multirow{2}{*}{Unrolling} & \multirow{2}{*}{$(8/7)$} & Final & 78.03 & 48.03 & 42.67 & \multirow{2}{*}{\textbf{35.4}}  & \multirow{2}{*}{-}\\
        ~  & ~ & Early & 79.57 & 54.01 & 51.90 & ~ & ~ \\        
    \bottomrule
    \end{tabular}}
\end{table*}

\begin{figure*}
    \centering
    \includegraphics[width=0.9\textwidth]{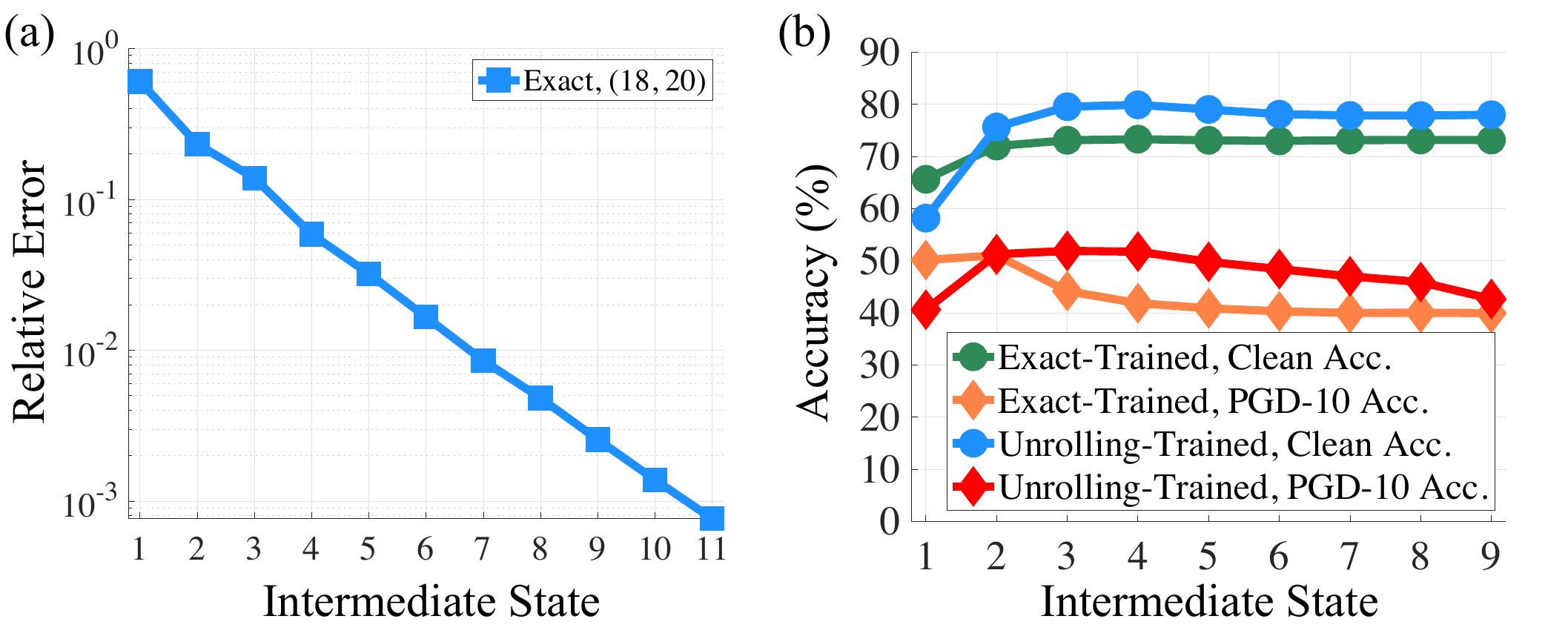}
    \caption{Performance of the (18/20) exact-trained DEQ-Large. (a) The relative error of each state $\bz_n$ in the forward pass of the exact-trained DEQ-Large under the $(18/20)$ setting. (b) Robustness accumulation effect in DEQ-Large: the (18/20) exact-trained DEQ v.s. the (8/7) exact-trained DEQ.}
    \label{fig:relerr-18-20-rebuttal}
\end{figure*}

From Table \ref{tab:exact-trained-iter-threshold}, it is witnessed that increasing the forward and the backward iterations imposes stabilization effect on DEQ training with the exact gradient. As a consequence, no false-positive robustness is observed in the settings of iteration pairs $(18/20)$. The relative error for each state $\bz_n$ in the forward pass under the $(18/20)$ setting is shown in Fig. \ref{fig:relerr-18-20-rebuttal}-(a). 

In this setting, the forward solver may require less number of iterations than the threshold $18$. Here we plot the relative errors of the first $11$ states in the forward solver. From Fig. \ref{fig:relerr-18-20-rebuttal}-(a) (also the blue line in Fig. \ref{fig:figure1-real}-(b)), the $(18/20)$ exact-trained DEQ does not violate its fixed-point structure, thus the gradient obfuscation issue is avoided. However, both of its training speed and robustness performance are outperformed by the unrolling-trained DEQ in the $(8/7)$ setting. 

\looseness=-1 Similar to Figures \ref{fig:train-exact-train-unroll-large} and \ref{fig:train-exact-train-unroll-xl}, we also compare the robustness accumulation effect between the $(18/20)$ exact-trained DEQ with the $(8/7)$ unrolling-trained DEQ. Figure \ref{fig:relerr-18-20-rebuttal}-(b) shows the comparison: clean and PGD-10 accuracies of both the final and the top intermediate states in the (18/20) exact-trained DEQ are lower than the (8/7) unrolling-trained DEQ. This is similar with the conclusion drawn from Table \ref{tab:train-unroll-train-exact-robust-acc-pgd}.

\subsection{Jacobian regularization weights during training} \label{app:jac-reg}

Another way to stabilize the adversarial training process is to impose stricter regularization on the DEQs. \cite{jacobian-deq} propose Jacobian regularization to stabilize the standard training of DEQs. In this section, we vary the weight $\gamma$ of the Jacobian regularization during DEQ training to its effect. In standard training, the weight is set as $0.4$. In this section, we sweep $\gamma$ over $\{0.4, 0.8, 1.2, 1.6, 2.0, 3.2 \}$ to train DEQ-Large with the exact gradient with 8 forward and 7 backward iterations (8/7).

Similar to Fig. \ref{fig:figure1-real}-(b), we plot the relative error at each $\bz_n$'s under different $\gamma$'s settings.

\begin{figure}[h]
    \centering
    \includegraphics[width=\textwidth]{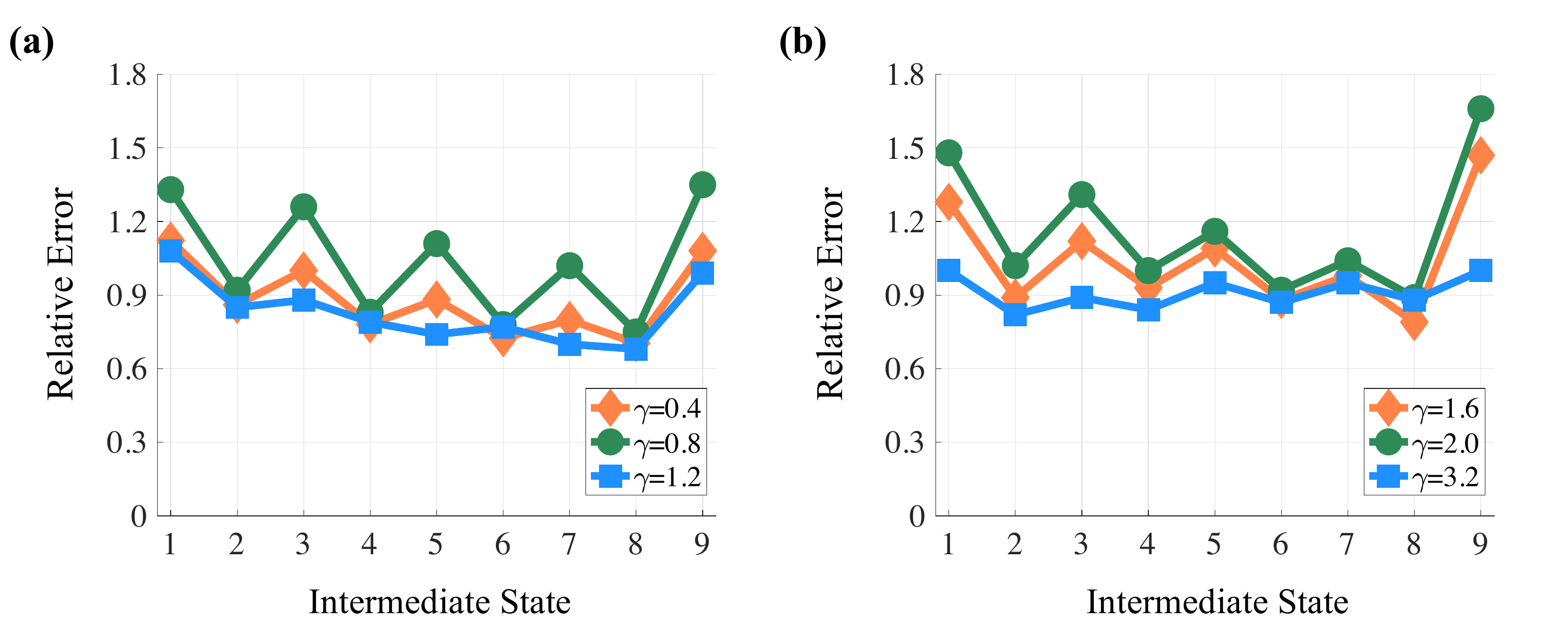}
    \caption{Relative errors at each intermediate state $\bz_n$ under different $\gamma$'s settings.}
    \label{fig:relerr_gm}
\end{figure}

\begin{figure}[h]
    \vspace{-10pt}
    \centering
    \includegraphics[width=\textwidth]{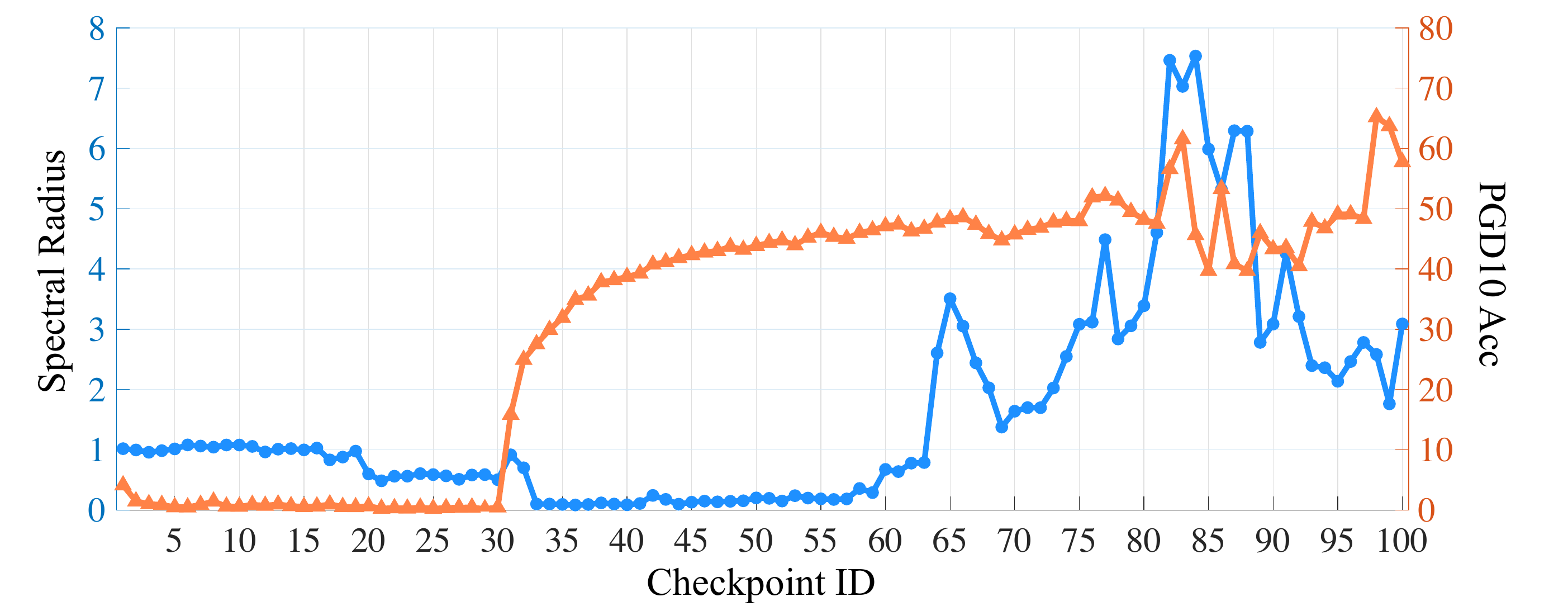}
    \caption{$\gamma=3.2$, (8/7) exact-trained DEQ-Large. The trace of spectral radius on the development set and the accuracy of each checkpoint under the ready-made PGD-10 attack. The blue line traces the spectral radius, and the orange line traces the accuracy under ready-made PGD-10. }
    \label{fig:sr_reg32}
    \vspace{-10pt}
\end{figure}

\begin{figure}[h]
    \centering
    \includegraphics[width=\textwidth]{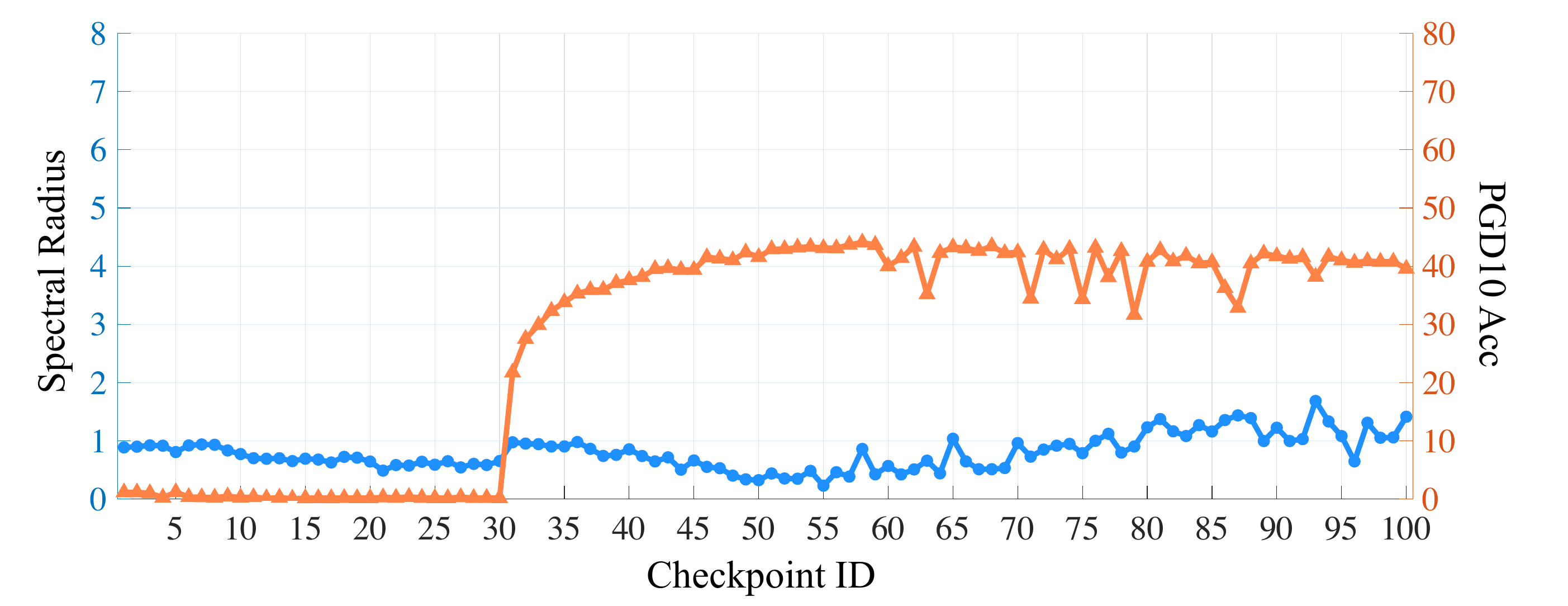}
    \caption{$\gamma=0.4$, (8/7) unrolling-trained DEQ-Large. The trace of spectral radius on the development set and the accuracy of each checkpoint under the ready-made PGD-10 attack. The blue line traces the spectral radius, and the orange line traces the accuracy under ready-made PGD-10.}
    \label{fig:sr_unroll}
    \vspace{-10pt}
\end{figure}

Illustrated in Fig. \ref{fig:relerr_gm}, all of the relative errors are larger than $0.6$, indicating the violation of the fixed-point structure. We have also explored the stability of the adversarial training process in the $\gamma=3.2$, (8/7) exact-trained setting and the $\gamma=0.4$, (8/7) unrolling-trained setting. We calculate the averaged spectral radius on the development set for all the checkpoints along the training, and plot them, together with their accuracy under the ready-made PGD-10 attack, in Figs \ref{fig:sr_reg32} and \ref{fig:sr_unroll}.

Shown in Fig. \ref{fig:sr_unroll}, the $\gamma=0.4$, (8/7) unrolling-trained DEQ-Large model always has a spectral radius less than or around $1.0$. In contrast, the spectral radius of the $\gamma=3.2$, (8/7) exact-trained DEQ-Large model becomes far larger than $1.0$ since the $65$-th checkpoint in Fig. \ref{fig:sr_reg32}. This coincides with the violated fixed-point structure, although achieving high ($>60$, but false-positive) accuracy under the ready-made PGD-10 attack. In this work, we use the checkpoint with the highest robustness under the ready-made PGD10 along the adversarial training process for our study. We leave the 
study of the white-box robustness evaluation for other checkpoints in future work.

When the fixed-point structure is broken, query-based attacks have a drastic effect on reducing the classification accuracy of DEQs (see Table \ref{tab:robustness-exact-grad}). We use SQUARE to attack these exact-trained DEQ-Large models with varied Jacobian regularization weight $\gamma$'s.

\begin{table}[h]
    \centering
    \vspace{-5pt}
    \caption{(8/7) exact-trained DEQ-Large with varied Jacobian regularization weight $\gamma$'s, in comparison with ($\gamma=0.4$): the (18/20) exact-trained DEQ-Large and the (8/7) unrolling-trained DEQ-Large.}
    \resizebox{\textwidth}{!}{\begin{tabular}{lcccccc|cl}
        \toprule
        $\gamma$ & 0.4 & 0.8 & 1.2 & 1.6 & 2.0 & 3.2 & 0.4 (18/20) & 0.4 (Unroll)\\
        \midrule
        Clean & 78.24 & 82.27 & 80.64 & 78.82 & 67.34 & 66.70 & 73.16 & 78.03 \\
        PGD (\textit{ready-made}) & 79.97 & 71.00 & 48.71 & 62.23 & 61.77 & 65.22 & 39.92 & 42.67 \\
        SQUARE & 5.95 & 10.00 & 32.54 & 23.85 & 4.21 & 2.72 & 47.20 & 45.34 \\
        \bottomrule
    \end{tabular}}
    \label{tab:sweep_gamma}
\end{table}

Table \ref{tab:sweep_gamma} compares the performance of the models with different $\gamma$'s under SQUARE and the ready-made PGD attack. For the (8/7) exact-trained DEQ-Large with varied $\gamma$'s, the SQUARE attack appears to be more powerful than the ready-made (gradient-based) PGD attack and always leads to severe robustness degradation. According to \cite{athalye2018obfuscated,carlini2019evaluating}, such a phenomenon indicates gradient obfuscation. For the exact-trained DEQ-Large with more iterations in the DEQ solver and the unrolling-trained DEQ-Large, as they have retained the fixed-point structure (Fig. \ref{fig:figure1-real}-(b) and Fig. \ref{fig:relerr-18-20-rebuttal}-(a)), we find the gradient-based PGD attack to be more effective. However, we emphasize that the two models \textit{still} suffer from gradient obfuscation: shown in Fig. \ref{fig:relerr-18-20-rebuttal}-(b), robustness accumulation effect is observed in both of the models (despite the retained fixed-point structure). The reason of the effect, as we have mentioned in Sec. \ref{sec:compare-exact-unrolling-under-white-attacks}, is that the black-box solver in DEQs results in misaligned gradients, which avoids the ready-made attacks to ``directly" attack the intermediate states. This has motivated us to propose intermediate/ensemble attacks and defenses for white-box robustness evaluation.

\begin{table}[h]
    \centering
    \vspace{-5pt}
    \caption{To summarize our work: the two types of gradient obfuscation that we have studied, and what we have done for our white-box robustness evaluation.}
    \resizebox{\textwidth}{!}{\begin{tabular}{l|l|l|l}
        \toprule
        State investigated & Phenomenon observed & The reason of the phenomenon & What we have done \\
        \midrule
        Final & SQUARE is more effective & Violated fixed-point structure & \makecell[l]{``Fix" the structure by: \\ Using the unrolling-trained DEQs; \\ Increasing the iterations in the solver; \\ Attempting with varied regularization weights.} \\
        \midrule
        Intermediate & Robustness accumulation  & Black-box solvers & Propose several white-box attacks and defenses. \\
        \bottomrule
    \end{tabular}}
    \label{tab:summarize}
\end{table}

To summarize, we have studied two types of gradient obfuscation in our work (see Table \ref{tab:summarize}). The violated fixed-point structure can be remedied by different techniques: we adopt unrolling-trained DEQs in Sec. \ref{sec:violation-of-fps}, increase the solver iterations in Sec. \ref{sec:compare-exact-unrolling-under-white-attacks} and Appendix \ref{appendix-f-2}, and attempt with stricter regularization in Appendix \ref{app:jac-reg}. We will explore with more regularization techniques in future work. However, the robustness accumulation effect \textit{always} exists. This ultimately urges the necessity of white-box robustness evaluation, and we propose several white-box attacks and defenses in this work. The two types of gradient obfuscation also echo with the two challenges in Sec. \ref{section-challenges}.


\newpage
\section{Memory and time complexity}

\subsection{On the O(1) memory concern of the proposed attacks and defenses}\label{appendix-g-memory-concern}

The unique property of DEQs lies in its O(1) memory consumption. It is therefore necessary to study whether the O(1) memory constraint still applies in the proposed attacks and defenses. 

\begin{wraptable}{r}{0.5\textwidth}
    \centering
    \caption{The memory usage of different defense strategies used in the (8/7) unrolling-trained DEQ-Large. No extra computation is needed. }    
    \begin{tabular}{c|c}
        \toprule
        Defense & Mem (GB) \\
        \midrule
        Final & 3.77 \\
        Early & 3.77 \\
        Ensemble & 3.77\\
        \bottomrule
    \end{tabular}
    \label{tab:mem_usage}
\end{wraptable}

From the attacking aspect, white-box attackers are assumed to have full access to the model, therefore they are allowed to ``open the black box" to trace all the intermediate steps in the solver. To fully evaluate the worst-case performance of models, white-box attackers are usually not constrained by the O(1) memory. From the defending aspect, as stated in Sec. \ref{section:white-box-defenses}, our defense methods still require only O(1) memory. For early state defense, we determine the optimal time to early exit the solver on the development set offline for once and then fix the early exit step during testing. For ensemble state defense, we maintain an accumulator ($\mathbf{\mathrm{ret}} \,+\!= \bz_n$) without storing the intermediate states of $\bz_n$ and output with $\mathbf{\mathrm{ret}}/N$ instead of $\bz_N$: Table \ref{tab:mem_usage} shows the empirical results on memory usage.

\subsection{On the time complexity concern of simultaneous adjoint}\label{appendix-k-time-complexity}

The core idea of the simultaneous adjoint is to \textbf{reuse} the approximated Jacobian inverse $B_n$ in the forward calculation of $\bz_n$ when calculating the adjoint state $\bu_n$. As a result, the "approximated Jacobian inverse" in the simultaneous adjoint does not need extra calculation. This is different from the original DEQ design where the forward and the backward passes are decoupled by separate fixed-point solvers, where different $B_n$'s need to be maintained separately.

Specifically, compared with the original forward pass (Eqs. (7) and (8)), the simultaneous adjoint calculation augments it with Eqs. (9) and (10). The time complexity of Eq.(10) is equilavent to Eq.(7). Eq. (9) calculates the residual of Eq.(3), the fixed-point equation of the backward pass, at $\bu^*$ = $\bu_n$ and $\bz^*$ = $\bz_n$. The complexity of this calculation is equivalent to just the \textbf{residual evaluation} when solving for the exact gradient in Eq.(3). 

In practice, the running time for the adaptive PGD-10 attack with the final adjoint state in "Simultaneous Adjoint" is 6,479ms per batch. In comparison, the running time for the adaptive PGD-10 attack with the unrolled final state in "Unrolled Intermediates" is 5,369ms per batch. It can be seen that the introduced computational burden of Eqs.(9) and (10) does not take the majority of the running time.

\section{Limitations and Broader Impact}

Adversarial attacks are fatal to deep learning methods, causing severe security risks in model deployment. To this end, reliable techniques are needed to defend against the attacks. In this work, we study the white-box robustness of general DEQ models. We observe the gradient obfuscation effect with ready-made attacks and propose several white-box attacks and defenses to facilitate white-box robustness evaluation. Our work contributes to the safety of general DEQs in white-box settings.

In this work, we did not test our methods on the adversarially-trained DEQs on ImageNet due to the time limit. Recent work \cite{mdeq,jacobian-deq} has shown that DEQ models work well on large-scale vision tasks, including ImageNet and Cityscapes. Given this, we could apply advanced adversarial training algorithms \cite{Wong2020Fast} to DEQs, which preserves the scalability of our methods. We also leave the white-box robustness evaluations on ImageNet as our future work. In addition, future work also includes efficient stabilization of the adversarial training procedure, as well as advanced white-box attacks and defense strategies for DEQs.

\end{document}